\documentclass[10pt,journal,compsoc]{IEEEtran}

\newcommand{\Tref}[1]{Table~\ref{#1}}

\newcommand{\Eref}[1]{Equation~\eqref{#1}}
\newcommand{\fref}[1]{Fig.~\ref{#1}}
\newcommand{\Fref}[1]{Figure~\ref{#1}}
\newcommand{\Frefs}[1]{Figures~\ref{#1}}
\newcommand{\sref}[1]{Sec.~\ref{#1}}

\usepackage{xcolor}
\newcounter{todos}
\AtEndDocument{\ifnum\value{todos}>0 \PackageWarning{TODOS}{There are \arabic{todos} todos left in this paper! Fix them before submitting the paper!} \fi}

\newcommand{\argmin}{\mathop{\mathrm{argmin}}\limits}

\newcommand{\V}[1]{\ensuremath{\mathbf{#1}}}

\ifCLASSINFOpdf
  \usepackage[pdftex]{graphicx}
\else
\fi

\usepackage{amsmath}
\usepackage{amssymb}
\usepackage{gensymb} %

\usepackage{booktabs}
\usepackage{colortbl}
\usepackage{makecell}
\newcommand*{\MinNumber}{0}
\newcommand*{\MidNumber}{12} 
\newcommand*{\MaxNumber}{40}
\newcommand*{\Ratio}{70}

\usepackage{pgfplots}
\usepgfplotslibrary{dateplot}
\newcommand{\cl}[1]{%
        \ifdim #1 pt > \MidNumber pt
            \pgfmathsetmacro{\PercentColor}{max(min(\Ratio*(#1 - \MidNumber)/(\MaxNumber-\MidNumber),\Ratio),0.00)} %
            \edef\x{\noexpand\cellcolor{red!\PercentColor!yellow!80}}\x #1
        \else
            \pgfmathsetmacro{\PercentColor}{max(min(\Ratio*(\MidNumber - #1)/(\MidNumber-\MinNumber),\Ratio),0.00)} %
            \edef\x{\noexpand\cellcolor{green!\PercentColor!yellow!80}}\x #1
        \fi
}

\usepackage{xspace}
\newcommand{\etal}{\textit{et~al.}}
\newcommand{\eg}{\textit{e}.\textit{g}.}
\newcommand{\ie}{\textit{i}.\textit{e}.}

\newcommand{\vn}{\boldsymbol{n}}
\newcommand{\vv}{\boldsymbol{v}}
\newcommand{\vl}{\boldsymbol{l}}

\newcommand{\B}{\textbf}

\usepackage{tikz}
\usepackage{tikz-3dplot}

\usepackage{subcaption}
\usepackage{url}
\usepackage[pagebackref=true,breaklinks=true,letterpaper=true,colorlinks,bookmarks=false]{hyperref}

\newcommand{\old}[1]{}
\newcommand{\rev}[1]{#1}
\newcommand{\revision}[1]{#1}
\renewcommand{\paragraph}[1]{\vspace{0.2em}\noindent \textbf{#1 \hspace{0.2em}}}

\begin{document}
\title{Deep Photometric Stereo for Non-Lambertian Surfaces}

\author{Guanying~Chen,~%
        Kai~Han,~%
        Boxin~Shi,~%
        Yasuyuki~Matsushita,~%
        and~Kwan-Yee~K.~Wong%
\IEEEcompsocitemizethanks{
\IEEEcompsocthanksitem G. Chen and K-Y. K. Wong are with The University of Hong Kong, Hong Kong, China. \protect\\E-mail: \{gychen,kykwong\}@cs.hku.hk
\IEEEcompsocthanksitem K. Han is with University of Oxford, Oxford, United Kingdom. \protect\\E-mail: khan@robots.ox.ac.uk
\IEEEcompsocthanksitem B. Shi is with Peking University, Beijing and Peng Cheng Laboratory, Shenzhen, China. \protect \\E-mail: shiboxin@pku.edu.cn
\IEEEcompsocthanksitem Y. Matsushita is with Osaka University, Osaka, Japan. \protect\\E-mail: yasumat@ist.osaka-u.ac.jp}%
\thanks{}}

\markboth{IEEE TRANSACTIONS ON PATTERN ANALYSIS AND MACHINE INTELLIGENCE}%
{Shell \MakeLowercase{\textit{et al.}}: Bare Demo of IEEEtran.cls for Computer Society Journals}

\IEEEtitleabstractindextext{%
\begin{abstract}
This paper addresses the problem of photometric stereo, in both calibrated and uncalibrated scenarios, for non-Lambertian surfaces based on deep learning. 
We first introduce a fully convolutional deep network for calibrated photometric stereo, which we call PS-FCN. Unlike traditional approaches that adopt simplified reflectance models to make the problem tractable, our method directly learns the mapping from reflectance observations to surface normal, and is able to handle surfaces with general and unknown isotropic reflectance.
At test time, PS-FCN takes an arbitrary number of images and their associated light directions as input and predicts a surface normal map of the scene in a fast feed-forward pass. 
To deal with the uncalibrated scenario where light directions are unknown, we introduce a new convolutional network, named LCNet, to estimate light directions from input images. The estimated light directions and the input images are then fed to PS-FCN to determine the surface normals. Our method does not require a pre-defined set of light directions and can handle multiple images in an order-agnostic manner. Thorough evaluation of our approach on both synthetic and real datasets shows that it outperforms state-of-the-art methods in both calibrated and uncalibrated scenarios.
\end{abstract}

}

\maketitle

\IEEEdisplaynontitleabstractindextext

\IEEEpeerreviewmaketitle

\IEEEraisesectionheading{\section{Introduction}\label{sec:introduction}}

\IEEEPARstart{P}{hotometric} stereo aims at recovering the surface normals of a static scene from a set of images captured under different light directions with a fixed camera~\cite{woodham1980ps,silver1980determining}. Based on the availability of calibrated lighting conditions, photometric stereo can be categorized into \emph{calibrated} and \emph{uncalibrated} photometric stereo settings.
Early calibrated photometric stereo methods assumed a simplified reflectance model, such as the ideal Lambertian model~\cite{woodham1980ps,silver1980determining} or analytical reflectance models~\cite{tozza2016direct,chung2008efficient,ruiters2009heightfield}. 
However, most of the real-world objects are non-Lambertian, and a specific analytical model is only valid for a small set of materials. A bidirectional reflectance distribution function (BRDF) is a general form for describing the reflectance property of a surface, but it is difficult to directly use a non-parametric form of BRDFs for photometric stereo. 

Recently, with the great success of deep learning in various computer vision tasks, deep learning based methods have been introduced to calibrated photometric stereo to handle surfaces with general and unknown isotropic reflectance~\cite{santo2017deep,ikehata2018cnn,Taniai18}. Instead of explicitly modeling complex surface reflectances, they directly learn the mapping from reflectance observations to surface normals given known light directions. 
However, the method in~\cite{santo2017deep} depends on a pre-defined set of light directions during training and testing. The methods in~\cite{santo2017deep,ikehata2018cnn} estimate the surface normals in a pixel-wise manner, making them not possible to account for the local context information of a surface point (\eg, surface smoothness prior).
Taniai and Maehara~\cite{Taniai18} introduced an optimization framework based on deep neural network, but their method suffers from complex scenes and requires a long processing time. 
Moreover, all of these methods assume known light directions.

On the other hand, the problem of uncalibrated photometric stereo still remains an open challenge, and a reliable method is desired for this relevant setting, because it eliminates the need for tedious light source calibration.
Most of the existing methods for uncalibrated photometric stereo~\cite{alldrin2007r,shi2010self,papad14closed} assume a simplified reflectance model, such as the Lambertian model, and focus on resolving the shape-light ambiguity, such as the Generalized Bas-Relief (GBR) ambiguity~\cite{belhumeur1999bas}. Although methods of~\cite{lu2013uncalibrated,lu2015uncalibrated} can handle surfaces with general BRDFs, they rely on the assumption of a uniform distribution of light directions for deriving a solution.

\begin{figure*}[t] \centering
    \includegraphics[width=0.96\textwidth]{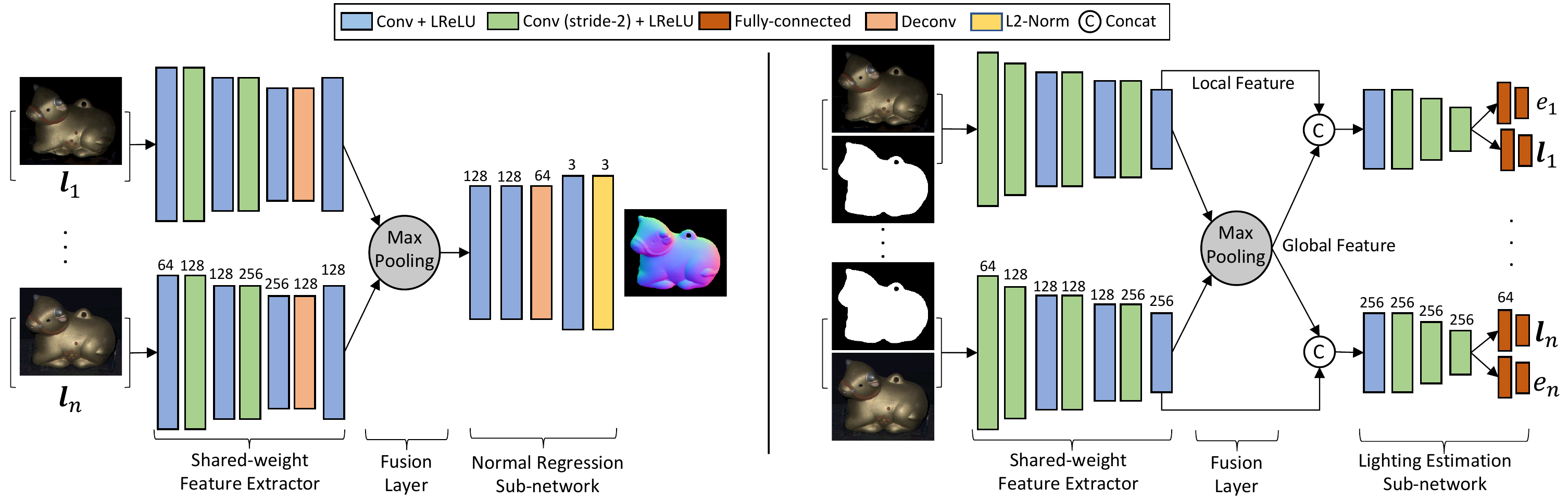} \\ \vspace{-0.3em}
    \makebox[0.48\textwidth]{\footnotesize (a) Network architecture of PS-FCN} 
    \makebox[0.48\textwidth]{\footnotesize (b) Network architecture of LCNet} 
    \caption{Overview of the proposed method. Values above the layers indicate the number of feature channels.} \label{fig:overview}
\end{figure*}

In this work, we study the problem of photometric stereo~(PS) for surfaces with general and unknown isotropic reflectance. 
\rev{Following the conventional practice, we assume an orthographic camera with a linear radiometric response, directional lightings coming from the upper-hemisphere range, and the viewing direction being parallel to the $z$-axis pointing towards the origin of the world coordinates.}

We first introduce a deep fully convolutional network (FCN), named PS-FCN, for \emph{calibrated photometric stereo}. PS-FCN takes an arbitrary number of images with their associated light directions as input, and predicts a surface normal map of the scene in a fast feed-forward pass (see~\fref{fig:overview}~(a)). 
Compared with previous learning based methods, our method does not depend on a pre-defined set of light directions during training and testing, and can handle multiple images in an order-agnostic manner. 
Moreover, convolutional neural network (CNN) can naturally incorporate information of the observations at neighboring pixels for computing feature maps, allowing our method to take advantage of local context information. 
To handle \emph{uncalibrated photometric stereo} where light directions are unknown, one may consider to directly learn the mapping from images to surface normals without taking the light directions as input. However, as will be shown in \sref{sec:exp}, the performance of such a na\"{\i}ve model lags far behind those which take both images and light directions as input. 
Instead, we introduce another CNN, named Lighting Calibration Network (LCNet), to estimate light directions from input images (see~\fref{fig:overview}~(b)). The estimated light directions and the input images can then be used by PS-FCN to estimate the surface normals. 

To simulate complex non-Lambertian surfaces that are close to real-world scenes for training, we create two large-scale synthetic datasets using shapes from the blobby shape dataset~\cite{johnson2011shape} and the sculpture shape dataset~\cite{wiles2017silnet}, and BRDFs from the MERL BRDF dataset~\cite{matusik2003merl}. Once trained on the synthetic data, we show that our method can generalize well on real datasets, such as the DiLiGenT benchmark~\cite{shi2018benchmark}.
Extensive experiments on both synthetic and real datasets show that our approach outperforms existing methods in both calibrated and uncalibrated photometric stereo settings, clearly demonstrating its effectiveness.

We have presented preliminarily results of this work in~\cite{chen2018ps,chen2019SDPS_Net}, and this paper extends them in several aspects. First, we extend PS-FCN to handle surfaces with spatially-varying BRDFs (SVBRDFs) by introducing a simple yet effective data normalization strategy. Second, we present a more detailed network analysis, experimental results, and discussion of how our method handles cast shadow. Third, we provide a comprehensive comparison between our method and the recent state-of-the-art methods.
\rev{Last, we discuss how LCNet resolves the ambiguity in lighting estimation and its limitation.} 
Our code, datasets, and models can be found at \url{https://guanyingc.github.io/SDPS-Net}.

\section{Related work}
\label{sec:related_work}
In this section, we review representative calibrated photometric stereo for non-Lambertian surfaces and uncalibrated photometric stereo methods. We also briefly review the loosely related work on learning based lighting estimation. Readers are referred to~\cite{shi2018benchmark,herbort2011introduction} for more comprehensive surveys of photometric stereo methods.

\rev{Consider a non-Lambertian surface whose appearance is described by a general isotropic BRDF $\rho$. Given a surface point with a unit surface normal vector $\vn \in \mathcal{S}^2$, $\mathcal{S}^2 = \{\vv \in \mathbb{R}^3 : \|\vv\|_2 = 1\}$ illuminated by the $j$-th incoming lighting with direction $\vl_j \in \mathcal{S}^2$ and intensity $e_j \in \mathbb{R}_+$, the image formation model from a fixed viewpoint can be written as
\begin{equation}
    \label{eq:formation}
    m_j = e_j \rho (\vn, \vl_j)~\text{max}(\vn^\top \vl_j, 0) + \epsilon_j,
\end{equation}
where $m$ is the measured intensity, $\text{max}(:, 0)$ accounts for attached shadows, and $\epsilon$ represents global illumination effects (\eg, cast shadows and inter-reflections) and noise.}

\paragraph{Calibrated photometric stereo}
\rev{For a Lambertian surface, the BRDF $\rho$ reduces to an unknown constant. Theoretically, 
the albedo scaled surface normal can be uniquely determined from the shadow-free observations captured under three non-coplanar light directions~\cite{woodham1980ps}.
However, perfect Lambertian surfaces barely exist. Many photometric stereo algorithms have been proposed to handle non-Lambertian surfaces.}
Outlier rejection based methods assume non-Lambertian observations to be local and sparse such that they can be treated as outliers. Various outlier rejection methods have been proposed so far. They are based on rank minimization~\cite{wu2010robust}, RANSAC~\cite{mukaigawa2007analysis}, taking median values~\cite{miyazaki2010median}, expectation maximization~\cite{wu2010photometric}, and sparse Bayesian regression~\cite{ikehata2012robust}. These outlier rejection methods generally require lots of input images and have difficulty in handling objects with non-sparse non-Lambertian observations (\eg, materials with broad and soft specular highlights).

\rev{Instead of rejecting specular observations as outliers, methods based on analytical reflectance models have been proposed. They adopt analytical models like Blinn-Phong model~\cite{tozza2016direct}, Ward model~\cite{chung2008efficient}, and Cook-Torrance model~\cite{ruiters2009heightfield}, to approximate the non-Lambertian reflectances.}
These methods require solving complex optimization problems, and can only handle limited classes of materials. Recently, bivariate BRDF representations~\cite{shi2014bi,ikehata2014p} were adopted to approximate isotropic BRDF, and a symmetry-based approach~\cite{holroyd2008photometric} was proposed to handle anisotropic reflectance without explicitly estimating a reflectance model. 

More recently, a few deep learning based methods have been introduced to calibrated photometric stereo~\cite{santo2017deep,Taniai18,ikehata2018cnn}. Santo~\etal~\cite{santo2017deep} proposed a fully-connected network to learn the mapping from reflectance observations captured under a pre-defined set of light directions to surface normal in a pixel-wise manner. 
Ikehata~\cite{ikehata2018cnn} introduced a fixed shape representation, called observation map, that is invariant to the number and permutation of the images. For each surface point of the object, all its observations are merged into an observation map based on the given light directions, and the observation map is then fed to a CNN to regress a normal vector. 
Compared with~\cite{santo2017deep,ikehata2018cnn}, our method can take advantage of local context information in predicting the surface normals, which results in a more robust behavior.
Taniai and Maehara~\cite{Taniai18} introduced an unsupervised learning framework that predicts both the surface normals and reflectance images of an object. Their model is ``trained'' at test time for each test object by minimizing the reconstruction loss between the input images and the rendered images, while our model is trained with supervised learning and achieves better performance on complex surfaces.

\paragraph{Uncalibrated photometric stereo}
\rev{Ignoring shadows and inter-reflections, the image formation model of a Lambertian surface is simplified to $m_j = e_j \rho \vn^\top \vl_j$. When light directions and intensities are unknown, the surface normals of a Lambertian object can only be estimated up to a $3\times 3$ linear ambiguity~\cite{hayakawa1994photometric}, given by 
\begin{equation}
    m_j = e_j\rho (\V{G}^{-\top} \vn)^\top (\V{G}\vl_j),\,\, \V{G} \in \mathbb{R}^{3\times 3}.
    \label{eq:gbr}
\end{equation}
This ambiguity can be reduced to a $3$-parameter GBR ambiguity using the surface integrability constraint, which also holds true at the presence of attached and cast shadows~\cite{belhumeur1999bas,yuille1999determining}.}
Previous work used additional clues like albedo priors~\cite{alldrin2007r,shi2010self}, inter-reflections~\cite{chandraker2005reflections}, specular spikes~\cite{drbohlav2005can}, Torrance and Sparrow reflectance model~\cite{georghiades2003incorporating}, reflectance symmetry~\cite{tan2007isotropy,wu2013calib}, multi-view images~\cite{esteban2008multiview}, and local diffuse maxima~\cite{papad14closed}, to resolve the GBR ambiguity. Cho~\etal~\cite{cho2016photometric} considered a semi-calibrated case where the light directions are known but not their intensities. There are a few works that can handle non-Lambertian surfaces under unknown lighting. Hertzmann and Seitz~\cite{hertzmann2005example} proposed an exemplar based method by inserting an additional reference object to the scene. Methods based on clues like similarity in radiance changes~\cite{sato2007shape,lu2013uncalibrated} and attached shadow~\cite{okabe2009attached} were also introduced, but they require the light sources to be uniformly distributed on the whole sphere. Recently, Lu~\etal~\cite{lu2018symps} introduced a method based on the ``constrained half-vector symmetry'' to work with non-uniform lightings. Different from these traditional methods, our method can deal with surfaces with general and unknown isotropic reflectance without the need of explicitly utilizing any additional clues or reference objects, solving a complex optimization problem at test time, or making assumptions on the light source distribution. 

\paragraph{Learning based lighting estimation}
Recently, learning based single-image lighting estimation methods have attracted considerable attention. Gardner~\etal~\cite{gardner2017learning} introduced a CNN for estimating HDR environment lighting from an indoor scene image. Hold-Goeffroy~\etal~\cite{hold2017deep} learned outdoor lighting using a physically-based sky model. Weber~\etal~\cite{weber2018learning} estimated indoor environment lighting from an image of an object with a known shape.  Zhou~\etal~\cite{Zhou_2018_CVPR} estimated lighting, in the form of Spherical Harmonics, from a human face image by assuming a Lambertian reflectance model. Different from the above methods, our method can estimate accurate directional lightings from multiple images of a static object with general shape and non-Lambertian surface.

\section{Learning photometric stereo}
In this section, we first introduce our strategy for adapting CNNs to handle a variable number of inputs, and then introduce a deep fully convolutional network, named PS-FCN, for learning calibrated photometric stereo. For learning uncalibrated photometric stereo, we introduce another CNN, named Lighting Calibration Network (LCNet), to estimate lightings from input images. LCNet can be seamlessly integrated with PS-FCN to predict accurate surface normals.
\rev{For the rest of this paper, we refer to light direction and intensity as ``lighting.''}

\subsection{Max-pooling for multi-feature fusion}
CNNs have been successfully applied to dense regression problems like depth estimation~\cite{eigen2014depth} and surface normal estimation~\cite{wang2015designing}, where the number of input images is fixed and identical during training and testing. Note that adapting CNNs to handle a variable number of inputs during testing is not straightforward, as convolutional layers require the input to have a fixed number of channels during training and testing. Given a variable number of inputs, a shared-weight feature extractor can be used to extract features from each of the inputs (\eg, siamese networks~\cite{bromley1993signature}), but an additional fusion layer is required to aggregate such features into a representation with a fixed number of channels. A convolutional layer is applicable for multi-feature fusion only when the number of inputs is fixed. Unfortunately, this is not practical for photometric stereo where the number of inputs often varies.

One possible way to tackle a variable number of inputs is to arrange the inputs sequentially and adopt a recurrent neural network (RNN) to fuse them. For example, Choy~\etal~\cite{choy20163d} introduced a RNN framework to unify single- and multi-image 3D voxel prediction. The memory mechanism of RNN enables it to handle sequential inputs, but at the same time also makes it sensitive to the order of inputs. This order sensitive characteristic is not desirable for photometric stereo as it will restrict the illumination changes to follow a specific pattern, making the model less general.

More recently, order-agnostic operations (\eg, pooling layers) have been exploited in CNNs to aggregate multi-image information. Wiles and Zisserman~\cite{wiles2017silnet} used max-pooling to fuse features of silhouettes from different views for novel view synthesis and 3D voxel prediction. Hartmann~\etal~\cite{hartmann2017learned} adopted average-pooling to aggregate features of multiple patches for learning multi-patch similarity. In general, max-pooling operation can extract the most salient information from all the features, while average-pooling can smooth out the salient and non-activated features. 

For photometric stereo, we argue that max-pooling is a better choice for aggregating features from multiple inputs. 
Our motivation is that, under a certain light direction, regions with high intensities or specular highlights provide strong clues for surface normal inference (\eg, for a surface point with a sharp specular highlight, its normal is close to the bisector of the viewing and light directions). Max-pooling can naturally aggregate such strong features from images captured under different light directions. Besides, max-pooling can ignore non-activated features during training, making it robust to cast shadow. As will be seen in \sref{sec:exp}, our experimental results do validate our arguments. We observe from experiments that each channel of the feature map fused by max-pooling is highly correlated to the response of the surface to a certain light direction. Strong responses in each channel are found in regions with surface normals having similar directions. The feature map can therefore be interpreted as a decomposition of the images under different light directions (see \fref{fig:res_visual}).

\subsection{PS-FCN for calibrated photometric stereo}
\paragraph{Network architecture}
PS-FCN is a multi-input-single-output (MISO) network consisting of three components, namely a shared-weight {\em feature extractor}, a {\em fusion layer}, and a {\em normal regression sub-network} (see \fref{fig:overview}~(a)). It can be trained and tested using an arbitrary number of images with their associated light directions as input\footnote{For calibrated photometric stereo, the input images are normalized by light intensities, and each light direction is represented by a unit $3$-vector.}.

For each light direction, we have a $3$-channel input image with the dimensions of $3 \times h \times w$, where $h$ and $w$ are the image height and width, respectively. Concatenating images taken under $q$ different lightings $\{\vl_1, ..., \vl_q\}$, we have the data with the dimensions of $q \times 3 \times h \times w$. In addition, we represent the light vectors $\{\vl_1, ..., \vl_q\}$ as $3$-channel images having the same spatial resolution as the input images, resulting in another $q \times 3 \times h \times w$ data. Putting them together, we finally have $q \times 6 \times h \times w$ dimensional inputs to our model.
We separately feed the image-light pairs to the shared-weight feature extractor to extract a feature map from each of the inputs, and apply a max-pooling operation in the fusion layer to aggregate these feature maps. Finally, the normal regression sub-network takes the fused feature map as input and estimates a normal map of the object.

\begin{figure}[t] \centering
    \input{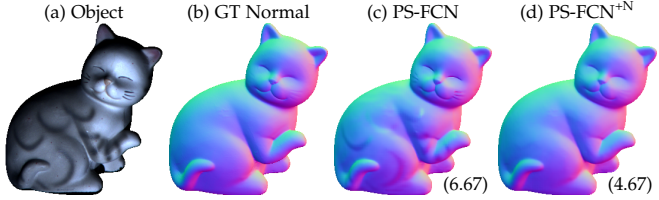}
    \caption{Comparison between PS-FCN and PS-FCN$^{+\text{N}}$ on an object with spatially-varying BRDFs. Numbers in the parentheses denote mean angular error (MAE) in degree.} \label{fig:cat_SVBRDF}
\end{figure}
The shared-weight feature extractor has seven convolutional layers, where the feature map is down-sampled twice and then up-sampled once, resulting in a down-sample factor of two. This design can increase the receptive field and preserve spatial information with a small memory consumption. 
The normal regression sub-network has four convolutional layers and up-samples the fused feature map to the same spatial dimension as the input images. An L2-normalization layer is appended at the end of the normal regression sub-network to produce the normal map.

As PS-FCN is a fully convolutional network, it can be applied to datasets with different image sizes. Thanks to the max-pooling operation in the fusion layer, PS-FCN possesses the order-agnostic property. 

\paragraph{Loss function}
Training of our PS-FCN is supervised by the estimation error between the predicted and the ground-truth normal maps. We formulate our loss function as the commonly used cosine similarity loss, given by
\begin{align}
    \label{eq:normal}
    \mathcal{L}_{\text{Normal}} = \frac{1}{hw} \sum_{i}^{hw} \left(1 - \vn_i^\top \tilde{\vn}_{i} \right),
\end{align}
where $\vn_{i}$ and $\tilde{\vn}_{i}$ denote the predicted normal and the ground-truth normal, respectively, at pixel $i$.
If the predicted normal has a similar orientation as the ground truth, the dot-product $\vn_{i} \cdot \tilde{\vn}_{i}$ will be close to $1$ and the loss becomes small, and vice versa. Other losses like mean squared error can also be alternatively adopted.

\paragraph{Extension to handle surfaces with SVBRDFs}
As PS-FCN is a fully-convolutional network that processes the input images in a patch-wise manner and is trained on surfaces with homogeneous BRDF, it may have difficulties in dealing with steep color changes caused by surfaces with SVBRDFs, as shown in~\fref{fig:cat_SVBRDF}~(c).
A straightforward idea to tackle this problem is to train a model on surfaces with SVBRDFs. However, creating a large-scale training dataset for this purpose is not trivial, since modeling surface appearance with realistic SVBRDFs requires manual editing from artists.
Even someone can collect a large-scale dataset of objects with SVBRDFs, the created dataset may not be able to faithfully cover the distribution of real data. 
In this paper, we introduce a simple yet effective data normalization strategy to enable PS-FCN to handle surfaces with SVBRDFs robustly. We will show that with the proposed data normalization strategy, our method can generalize well to surfaces with SVBRDFs after training only on surfaces with homogeneous BRDF.

\begin{figure}[t] \centering
    \input{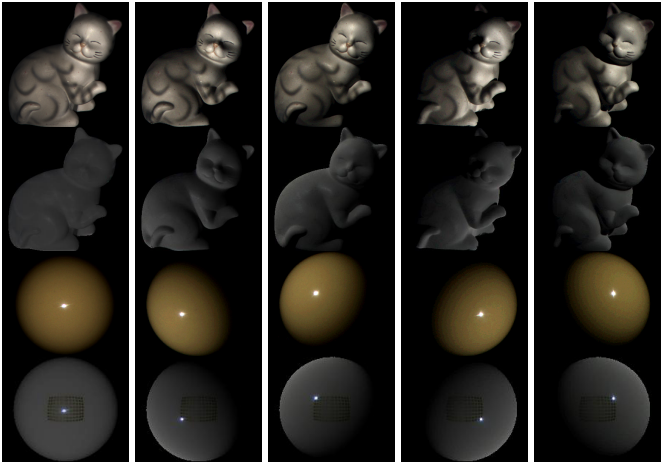}
    \caption{Illustration of the introduced data normalization operation on {\sc cat} and {\sc ball} in the DiLiGenT benchmark. The first and third rows show the original images, while the second and last rows show the normalized images. Only $5$ out of $96$ images for each object are shown.} \label{fig:normalization_SVBRDF}
\end{figure}

During training, given $q$ observations of a surface point\footnote{Note that the observations are already normalized by the light intensities.}, we concatenate all the observations and normalize them to a unit length vector by
\begin{align}
    \label{eq:normalize}
    \left(m_1', ..., m_{q}'\right) = \left(\frac{m_1}{\sqrt{m_1^2+...+m_{q}^2}}, ..., \frac{m_{q}}{\sqrt{m_1^2+...+m_{q}^2}}\right),
\end{align}
where $m$ and $m'$ represent the original and normalized observations, respectively (for RGB images, we perform normalization on each channel separately). 
The intuition behind this operation is as follows. Consider a Lambertian model, the BRDF $\rho(\vn, \vl)$ degenerates to a constant albedo $\rho$ and $m = \rho \max(\vn^\top \vl_j, 0)$. 
After the data normalization operation, we have
\begin{align}
    \label{eq:normalize2}
    m_{i}' = \frac{\max(\vn^\top \vl_i, 0)}{\sqrt{\max(\vn^\top \vl_1, 0)^2 + \cdots + \max(\vn^\top \vl_q, 0)^2}}.
\end{align}
\Eref{eq:normalize2} shows that the effect of albedo in Lambertian surfaces can be removed after performing data normalization, as shown in the first example in \fref{fig:normalization_SVBRDF}.

However, the above conclusion is not true for non-Lambertian surfaces, because for regions with specular highlights under some light directions, the observations under other light directions will be suppressed after data normalization (see the example of {\sc ball} in \fref{fig:normalization_SVBRDF}).
Nevertheless, we experimentally found that such a normalization strategy works equally well for non-Lambertian surfaces under the PS-FCN framework.
This might be explained by the fact that for a non-Lambertian surface under directional lighting, the low-frequency observations are quite close to Lambertian reflectance~\cite{shi2014bi}.
For observations exhibiting specular highlights under some light directions, the max-pooling operation in the fusion layer can naturally ignore the non-activated features (\ie, features extracted from the suppressed observations) and aggregate the most salient features.  
Note that this normalization strategy has also been adopted in~\cite{sato2007shape,lu2013uncalibrated} to compute the similarity between two pixel intensity profiles of non-Lambertian surfaces, while we use this normalization strategy as a preprocessing for CNNs to handle surfaces with SVBRDFs.  

When the number of input images at test time $t$ is different from that in training $q$, the magnitude of the normalized observations will be different, which leads to decreased performance (\eg, when all observations have the same values, 
we have $m_{\text{train}}' = 1/\sqrt{q}, m'_{\text{test}}=1/\sqrt{t}$). 
We experimentally verified that multiplying the normalized observations with the scalar $\sqrt{t/q}$ at test time solves this problem.
We trained a variant model of PS-FCN, denoted as PS-FCN$^\text{+N}$, using the proposed data normalization strategy.
\Fref{fig:cat_SVBRDF}~(d) shows an example result that PS-FCN$^\text{+N}$ performed better than PS-FCN on surfaces with SVBRDFs.

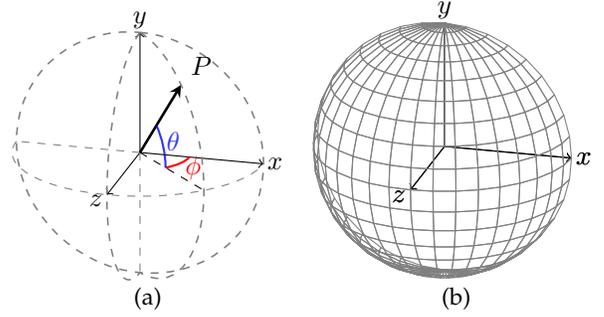
\begin{figure} \centering
    \tdplotsetmaincoords{70}{105}
\pgfmathsetmacro{\rvec}{1}
\pgfmathsetmacro{\thetavec}{40}
\pgfmathsetmacro{\phivec}{45}
\begin{tikzpicture}[scale=1.7,tdplot_main_coords]
    \coordinate (O) at (0,0,0); %
    \draw[->] (0,0,0) -- (1,0,0) node[below=0.5ex,left=-0.4ex]{$z$};
    \draw[->] (0,0,0) -- (0,1,0) node[right=-0.4ex]{$x$};
    \draw[->] (0,0,0) -- (0,0,1) node[above=-0.4ex]{$y$};
    \draw[dashed,color=gray] (0,0,0) -- (0,0,-1);
    \draw[dashed,color=gray] (0,0,0) -- (0,-1,0);
    \tdplotdrawarc[dashed,color=gray]{(O)}{1}{-90}{90}{}{}
    \tdplotsetthetaplanecoords{0}
    \tdplotdrawarc[semithick,dashed,color=gray,tdplot_rotated_coords]{(0,0,0)}{1}{0}{180}{}{}
    \tdplotsetthetaplanecoords{90}
    \tdplotdrawarc[semithick,dashed,color=gray,tdplot_rotated_coords]{(0,0,0)}{1}{0}{360}{}{}

    \tdplotsetcoord{P}{\rvec}{\thetavec}{\phivec} %
    \draw[line width=1pt,-stealth,color=black] (O) -- (P) node[above right] {$P$};

    \tdplotsetcoord{Px}{\rvec}{90}{\phivec} %
    \draw[dashed, color=black] (O) -- (Px);
    \tdplotdrawarc[thick,color=red]{(O)}{0.4}{\phivec}{90}{color=red,below=0.4ex,right=-0.4ex}{$\phi$} 

    \tdplotsetthetaplanecoords{\phivec} %
    \tdplotdrawarc[tdplot_rotated_coords,thick,color=blue,opacity=0.8]{(0,0,0)}{0.4}{\thetavec}{90}{above=0.4ex,right=-0.5ex}{$\theta$}
    \tdplotdrawarc[semithick,dashed,color=gray,tdplot_rotated_coords]{(0,0,0)}{1}{0}{180}{}{}
\end{tikzpicture}
    \raisebox{0.05\height}{\includegraphics[width=0.21\textwidth]{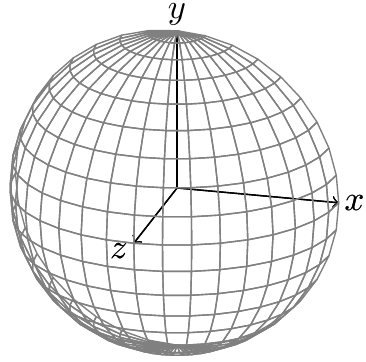}} \\ 
    \vspace{-0.9em} \makebox[0.23\textwidth]{\small (a)} \makebox[0.21\textwidth]{\small (b)} \\
    \caption{(a) Illustration of the coordinate system ($z$ axis is the viewing direction). $\phi \in [0\degree, 180\degree]$ and $\theta \in [-90\degree, 90\degree]$ are the azimuth and elevation of the light direction, respectively. (b) Example discretization of the light direction space when $K_d=18$.} \label{fig:coord}
\end{figure}

\subsection{LCNet for lighting estimation}
So far, we have assumed that the light intensities and directions are known. However, this assumption does not always hold in real applications. PS-FCN can be extended to handle the case where light directions are unknown by simply removing the light directions during training. However, as will be shown in \sref{sec:exp}, such a model is not optimal. To handle uncalibrated photometric stereo where both light intensities and directions are unknown, a more preferable solution is to learn the lightings from the input images, and then take the estimated lightings as part of the input for PS-FCN to estimate accurate normals.
To estimate lightings from the images, one straightforward idea would be to directly regressing the light direction vectors and intensity values, but we propose that formulating the lighting estimation as a classification problem is a superior choice, as will be verified by our experiments in \sref{sec:exp}. Our arguments are as follows. First, classifying a light direction into a certain range is easier than regressing the exact value(s), and this will reduce the learning difficulty. 
\rev{Second, taking discretized light directions as input allows PS-FCN to better tolerate small errors in the estimated light directions, as verified by the experimental results.}

\paragraph{Discretization of lighting space} Since we cast our lighting estimation as a classification problem, we need to discretize the continuous lighting space. Note that a light direction in the upper-hemisphere can be described by its azimuth $\phi \in [0\degree, 180\degree]$ and elevation $\theta \in [-90\degree, 90\degree]$ (see \fref{fig:coord}~(a)). We can discretize the light direction space by evenly dividing both the azimuth and elevation into $K_d$ bins, resulting in $K_d^2$ classes (see \fref{fig:coord}~(b)). Solving a $K_d^2$-class classification problem is not computationally efficient, as the softmax probability vector will have a very high dimension even when $K_d$ is not large (\eg, $K_d^2=1,296$ when $K_d=36$). 
Instead, we estimate the azimuth and elevation of a light direction separately, leading to two $K_d$-class classification problems. Similarly, we evenly divide the range of possible light intensities into $K_e$ classes (\eg, $K_e=20$ for a possible light intensity range of $[0.2, 2.0]$).

\paragraph{Local-global feature fusion}
A straightforward approach to estimate the lighting for each image would be simply taking a single image as input, encoding it into a feature map using a CNN, and feeding the feature map to a lighting prediction layer. It is not surprising that the result of such a simple solution is far from satisfactory. Note that the appearance of an object is determined by its surface geometry, reflectance model and the lighting. The feature map extracted from a single observation obviously does not provide sufficient information for resolving the shape-light ambiguity. Thanks to the nature of photometric stereo where multiple observations of an object are considered, we propose a local-global feature fusion strategy to extract more comprehensive information from multiple observations.

Specifically, we separately feed each image into a shared-weight feature extractor to extract a feature map, which we call \emph{local feature} as it only provides information from a single observation. All local features of the input images are then aggregated into a \emph{global feature} through a max-pooling operation.
Such a global feature is expected to convey implicit surface geometry and reflectance information of the object which help to resolve the ambiguity in lighting estimation. Each local feature is concatenated with the global feature, and fed to a shared-weight lighting estimation sub-network to predict the lighting for each individual image. By taking both local and global features into account, our model can produce much more reliable results than using local features alone. 
\rev{We also include the object mask as input, as it allows the network to focus on extracting useful features inside the object region.}

\paragraph{Network architecture}
LCNet is a multi-input-multi-output (MIMO) network that consists of a shared-weight \emph{feature extractor}, a \emph{fusion layer} (\ie, max-pooling layer), and a shared-weight \emph{lighting estimation sub-network} (see \fref{fig:overview}~(b)). 
It takes the observations of the object together with the object mask as input, and outputs the light directions and intensities in the form of softmax probability vectors of dimension $K_d$ (azimuth), $K_d$ (elevation) and $K_e$ (intensity), respectively. 
We convert the output of LCNet to $3$-vector light directions and scalar intensity values by simply taking the middle value of the range with the highest probability\footnote{We have experimentally verified that alternative ways like taking the expectation of the probability vector or performing quadratic interpolation in the neighborhood of the peak value do not improve the result.}.

\paragraph{Loss function}
Multi-class cross-entropy loss is adopted for both light direction and intensity estimation, and the overall loss function is
\begin{align}
    \label{eq:cls_loss}
    \mathcal{L}_{\text{Light}} & = \lambda_{l_a} \mathcal{L}_{l_a} + \lambda_{l_e} \mathcal{L}_{l_e} + \lambda_e \mathcal{L}_e,
\end{align}
where $\mathcal{L}_{l_a}$ and $\mathcal{L}_{l_e}$ are the loss terms for azimuth and elevation of the light direction, and  $\mathcal{L}_e$ is the loss term for light intensity. During training, weights $\lambda_{l_a}$, $\lambda_{l_e}$ and $\lambda_e$ for the loss terms are set to $1$.

\paragraph{Integration with PS-FCN} 
Given the light directions and intensities estimated by LCNet, PS-FCN can be directly applied to estimate the surface normals of an object. 
For uncalibrated photometric stereo, we found that training PS-FCN from scratch with the estimated lighting of LCNet instead of the ground-truth lighting can lead to a more robust behavior over noise in the lighting.

\section{Dataset}
The training of our models requires ground-truth normal maps of the objects. However, obtaining ground-truth normal maps of real objects is a difficult and time-consuming task. Hence, we create two synthetic datasets for training and one synthetic dataset for testing. The publicly available real photometric stereo datasets are reserved to validate the generalization ability of our models. 

\subsection{Synthetic data for training}
We used shapes from two existing 3D datasets, namely the blobby shape dataset \cite{johnson2011shape} and the sculpture shape dataset \cite{wiles2017silnet}, to generate our training data using the physically based raytracer Mitsuba \cite{jakob2010mitsuba}. Following SS17~\cite{santo2017deep}, we employed the MERL dataset \cite{matusik2003merl}, which contains $100$ different BRDFs of real-world materials, to define a diverse set of surface materials for rendering these shapes. 
Note that our datasets explicitly consider cast shadows during rendering.

\begin{figure}[t] \centering
    \input{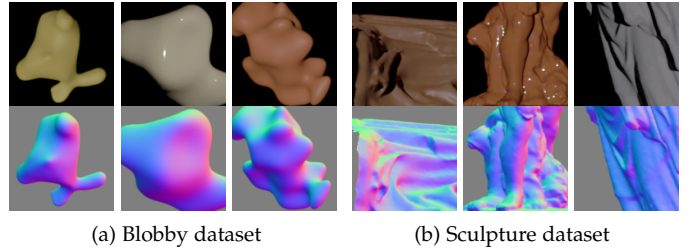}
    \caption{Examples of the synthetic training data (images are adjusted with gamma correction for visualization purpose).} \label{fig:data_samples}
\end{figure}

\paragraph{Blobby dataset} We first followed \cite{santo2017deep} to render our training data using the blobby shape dataset \cite{johnson2011shape}, which contains $10$ blobby shapes with various normal distributions. For each blobby shape, $1,296$ regularly-sampled views ($36$ azimuth angles $\times$ $36$ elevation angles) were used, and for each view, $2$ out of $100$ BRDFs were randomly selected, leading to $25,920$ samples ($10\times 36\times 36\times 2$).
For each sample, we rendered $64$ images with a spatial resolution of $128 \times 128$ under light directions randomly sampled from a range of 180\degree $\times$ 180\degree, which is more general than the range ($74.6\degree \times 51.4\degree$) used in the real data benchmark \cite{shi2018benchmark}. We randomly split this dataset into $99:1$ for training and validation (see \fref{fig:data_samples}~(a)).

\paragraph{Sculpture dataset} The surfaces in the blobby shape dataset are usually largely smooth and lack of details. To provide more complex (realistic) normal distributions for training, we employed $8$ complicated 3D models from the sculpture shape dataset introduced in~\cite{wiles2017silnet}. We generated samples for the sculpture dataset in exactly the same way we did for the blobby shape dataset, except that we discarded views containing holes or showing uniform normals (\eg, flat facets). 
The rendered images are with a size of $512 \times 512$ when a whole sculpture shape is in the field of view.
We then regularly cropped patches of size $128 \times 128$ from the rendered images and discarded those with a foreground ratio less than $50\%$.
This gave us a dataset of $59,292$ samples, where each sample contains $64$ images rendered under different light directions. Finally, we randomly split this dataset into $99:1$ for training and validation (see \fref{fig:data_samples}~(b)).

\begin{figure}[t] \centering
    \input{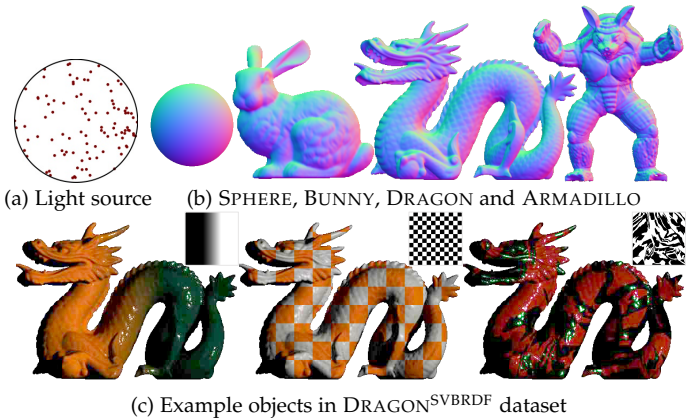}
    \caption{(a) Lighting distribution of SynTest$^\text{MERL}$ dataset. The light direction is visualized by mapping a $3$-d vector $[x,y,z]$ to a point $[x,y]$. (b) Ground-truth normals of {\sc Sphere}, {\sc Bunny}, {\sc Dragon}, and {\sc Armadillo}. (c) Visualization of the selected material maps (Ramp, Checker, Irregular) and examples in {\sc Dragon$^\text{SVBRDF}$} dataset.} \label{fig:syn_test_sample}%
\end{figure}

\paragraph{Training Details} %
During training, we applied noise perturbation in the range of $[-0.025, 0.025]$ for data augmentation.
To train PS-FCN for calibrated photometric stereo, given an image of size $128\times 128$, we randomly performed image rescaling (with the rescaled width and height within the range of $[32, 128]$, without preserving the original aspect ratio). Image patches of size $32\times 32$ were then randomly cropped for training. At test time, PS-FCN can take images of different sizes as input.

As the training data is rendered with uniform light intensity, to train LCNet for uncalibrated photometric stereo, we simulate images under different light intensities by randomly generated light intensities in the range of $[0.2, 2.0]$ to scale the magnitude of the images (\ie, the ratio of the highest light intensity to the lowest one is $10$)\footnote{Note that the ratio (other than the exact value) matters, since light intensity can only be estimated up to a scale factor.}. Note that this selected range contains a wider range of intensity values than the public photometric stereo datasets like DiLiGenT benchmark~\cite{shi2018benchmark} and Gourd\&Apple dataset~\cite{alldrin2008p}. 
As LCNet contains fully-connected layers and requires the input to have a fixed spatial dimension, the input image size for LCNet during both training and testing is $128\times 128$.

\subsection{Synthetic data for analysis}
To quantitatively evaluate the performance of our method on different materials and shapes, we rendered a synthetic test dataset including Sphere, Bunny, Dragon, and Armadillo shapes. Hereafter, we denote this test dataset as SynTest$^{\text{MERL}}$ and these shapes as {\sc Sphere}, {\sc Bunny}, {\sc Dragon}, {\sc Armadillo} respectively. 
Each shape was rendered with $100$ isotropic BRDFs from MERL dataset~\cite{matusik2003merl} under $100$ light directions randomly sampled from the upper-hemisphere, leading to $400$ test objects (see \fref{fig:syn_test_sample}~(a)-(b)).
Cast shadows and inter-reflections were considered during rendering using the physically based raytracer Mitsuba~\cite{jakob2010mitsuba}. 

To analyze how surfaces with SVBRDFs affect the performance of our method, we created another synthetic test dataset with SVBRDFs, denoted as {\sc dragon}$^{\text{SVBRDF}}$, following~\cite{goldman2010shape}.
Specifically, we blended two BRDFs from $100$ MERL dataset for {\sc dragon} using $3$ materials maps, namely the \emph{Ramp}, \emph{Checker}, and \emph{Irregular}, as shown in \fref{fig:syn_test_sample}~(c). Note that for each material map, there are $C(100,2)=4,950$ combinations of two BRDFs, leading to $14,850$ test objects.

\begin{figure}[t] \centering
    \input{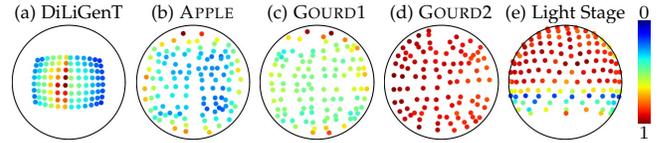}
    \caption{Lighting distributions of the real testing datasets. The color of the point indicates the light intensity (value is divided by the highest intensity to normalize to $[0,1]$).} \label{fig:light_dist}
\end{figure}

\begin{figure*}[t]%
    \centering
    \input{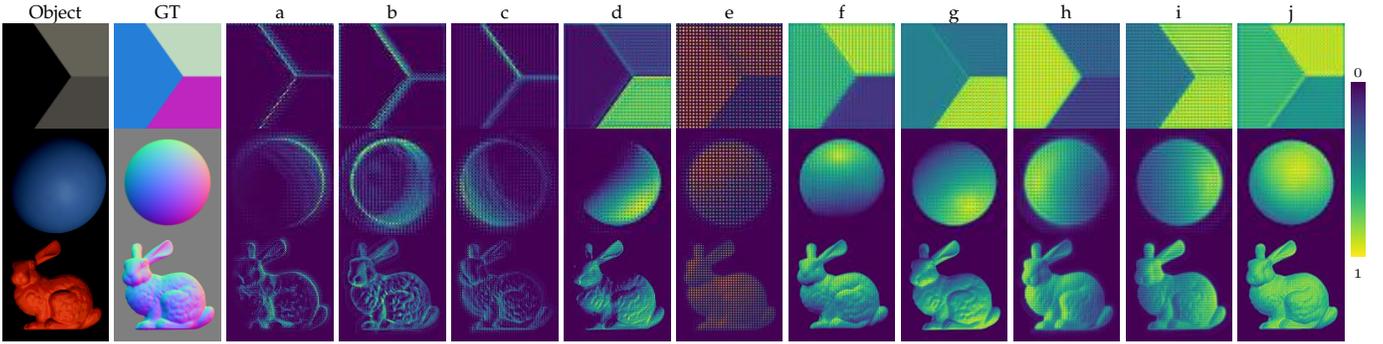}
    \caption{Visualization of the learned feature map after fusion (the features were normalized to $[0, 1]$). The first two columns show the objects and ground-truth normals. The subsequent columns (a-j) visualize $10$ out of $128$ channels of the fused feature map. Note that different regions with similar normal directions are fired in different channels. Each channel can therefore be interpreted as the probability of the normal belonging to a certain direction (or alternatively as the object shading rendered under a certain light direction). } \label{fig:res_visual}
\end{figure*}

\subsection{Real data for testing}
We employed three challenging real non-Lambertian photometric stereo datasets for testing, namely the \emph{DiLiGenT benchmark}~\cite{shi2018benchmark}, \emph{Gourd\&Apple dataset}~\cite{alldrin2008p}, and \emph{Light Stage Data Gallery}~\cite{einarsson2006relighting}. Note that none of these datasets were used in the training.

DiLiGenT benchmark~\cite{shi2018benchmark} is a public dataset containing $10$ real objects, and each object was captured under $96$ predefined light directions (see \fref{fig:light_dist}~(a)).  Both ground-truth lighting conditions and normal maps are provided. 
We quantitatively evaluated the performance of our method on both lighting and normal estimation.

Gourd\&Apple dataset~\cite{alldrin2008p} consists of three objects, namely {\sc Apple}, {\sc Gourd1}, and {\sc Gourd2}, with $112$, $102$ and $98$ images, respectively. 
\Frefs{fig:light_dist}~(b)-(d) visualize the lighting distributions of this dataset.  
Light Stage Data Gallery~\cite{einarsson2006relighting} is composed of six objects, and $253$ images are provided for each object. We only used $133$ images with the front side of the object under illumination. \Fref{fig:light_dist}~(e) visualizes the lighting distribution of the selected images.
Since these two datasets only provide calibrated lightings (without ground-truth normal maps), we quantitatively evaluated our method on lighting estimation but only qualitatively evaluated it on normal estimation.

\section{Experimental Results}
\label{sec:exp}
In this section, we present network analysis for our method, and compare our method with the previous state-of-the-art methods on both synthetic and real datasets.

\paragraph{Implementation details}
Our framework was implemented in PyTorch~\cite{paszke2017pytorch} and Adam optimizer~\cite{kingma2014adam} was used with default parameters ($\beta_1=0.9$ and $\beta_2=0.999$).
PS-FCN contains $2.2$ million learnable parameters.
We trained PS-FCN using a batch size of $32$ for $30$ epochs, and it only took a few hours for training to converge using a single NVIDIA Titan X Pascal GPU (\eg, about $9$ hours using $32$ image-light pairs per sample on both the blobby and sculpture datasets). Learning rate was initially set to $0.001$ and divided by $2$ every $5$ epochs. 

LCNet contains $4.4$ million parameters. We trained LCNet using a batch size of $32$ for $20$ epochs, and it took about $22$ hours to train LCNet on a single GPU with a fixed input image number of $32$. The learning rate was initially set to $0.0005$ and halved every $5$ epochs. 

\begin{table}[t] \centering
    \caption{Normal estimation results on SynTest$^\text{MERL}$ dataset. The results are averaged over samples rendered with $100$ BRDFs. B and S stand for the blobby and sculpture training datasets, respectively.} %
    \resizebox{0.49\textwidth}{!}{
    \Large
    \begin{tabular}{*{5}{c}|*{4}{c}}
            \toprule
            \multicolumn{5}{c}{Model Variants} & \multicolumn{4}{c}{Test Objects} \\
            ID & Data & Fusion &  Train \# & Test \# & {\sc Sphere} & {\sc Bunny} & {\sc Dragon} & {\sc Armod.} \\
            \midrule
            A0 & B   & Conv  & 32 & 32  & 4.54 & 6.74 & 9.57 & 9.87 \\
            A1 & B   & Max-p & 32 & 32  & 3.65 & 5.33 & 7.86 & 8.09 \\
            \midrule               
            A2 & B   & Avg-p & 32 & 100 & 3.71 & 5.36 & 8.17 & 7.92 \\
            A3 & B   & Max-p & 32 & 100 & 3.40 & 4.80 & 7.23 & 7.21 \\
            \midrule               
            A4 & B+S & Max-p & 32 & 100 & \B{2.66} & \B{3.80} & \B{4.83} & \B{5.24} \\
            \bottomrule
        \end{tabular}
}
 \label{tab:quant_calib_normal_syn}
\end{table}

\subsection{Results on calibrated photometric stereo}
To measure the accuracy of the predicted normal maps, mean angular error (MAE) in degree was used.

\subsubsection{Network analysis of PS-FCN with synthetic data}
We quantitatively analyzed PS-FCN on the synthetic dataset. 
For all the experiments in network analysis, we performed $100$ random trials (save for the experiments using all $100$ image-light pairs per sample during testing) and reported the average results.

\begin{figure}[t] \centering
    \includegraphics[width=0.24\textwidth]{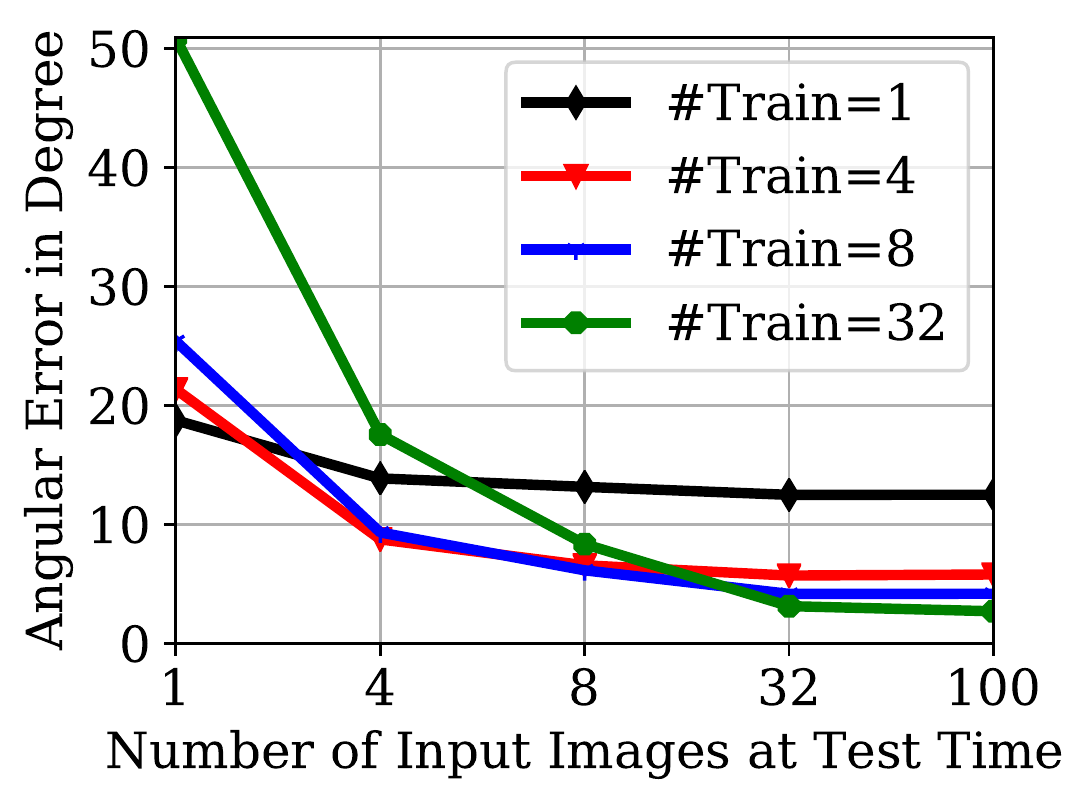}
    \includegraphics[width=0.24\textwidth]{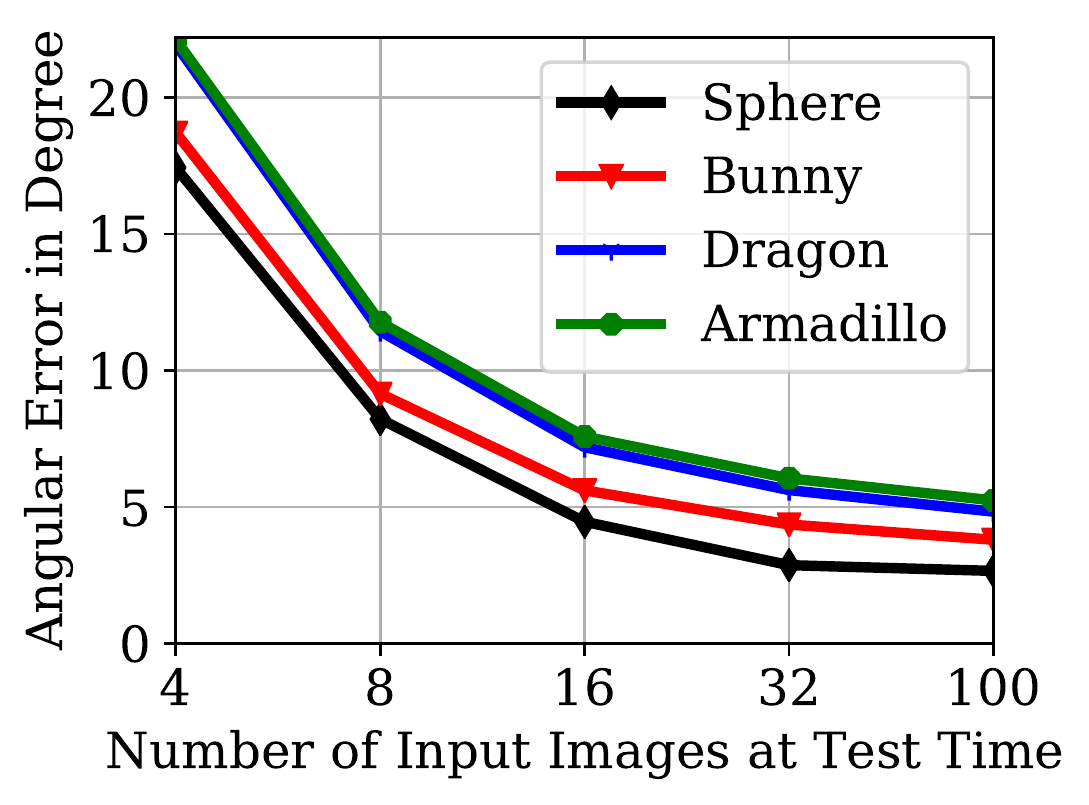} \\ \vspace{-0.5em}
    \makebox[0.24\textwidth]{\scriptsize (a)} \hfill
    \makebox[0.24\textwidth]{\scriptsize (b)}\\
    \caption{(a) Results of PS-FCN trained and tested with different numbers of input images on {\sc Sphere}. (b) Results of PS-FCN trained with a fixed number of $32$ input images and tested with different numbers of input images.} \label{fig:calib_img_num_syn}
\end{figure}

\paragraph{Effectiveness of max-pooling} We first validated the effectiveness of max-pooling in multi-feature fusion by comparing it with convolutional layers and average-pooling. 
Experiments with IDs A0 \& A1 in \Tref{tab:quant_calib_normal_syn} show that fusion by convolutional layer on the concatenated features was sub-optimal. This could be explained by the fact that the weights of the convolutional layer are related to the order of the input features, while the order of the input image-light pairs is random in our case, thus increasing the difficulty for the convolutional layer to find the relations among multiple features. 
Experiments with IDs A2 \& A3 compared the performance of average-pooling and max-pooling for multi-feature fusion. It can be seen that max-pooling performs consistently better than average-pooling on SynTest$^\text{MERL}$ dataset.
\Fref{fig:res_visual} visualizes the fused features by max-pooling for three objects with different shapes and reflectances. We can see that each channel of the fused features can be interpreted as the probability of the normal belonging to a certain direction, and max-pooling can naturally aggregate such information from multiple observations.

\paragraph{Effects of training data and input image number} 
By comparing experiments with ID A3 \& A4 in \Tref{tab:quant_calib_normal_syn}, we can see that training with the additional sculpture dataset that has a more complex normal distribution helped to boost the performance of PS-FCN. This result suggests that the performance of PS-FCN could be further improved by introducing more complex and realistic training data. 

\Fref{fig:calib_img_num_syn}~(a) shows that for a fixed number of inputs during testing, PS-FCN performs better when the number of inputs during training is close to that during testing. 
\rev{It is worth noting that when there is only one input image, the problem reduces to the more challenging shape-from-shading problem. \Fref{fig:calib_img_num_syn}~(a) shows that PS-FCN performs best when the training image number is also $1$, with an average MAE of $18.75\degree$ for {\sc Sphere}. However, this result is moderately inaccurate, indicating that PS-FCN has difficulties in resolving the ambiguity in the problem of shape from shading.}

\Fref{fig:calib_img_num_syn}~(b) shows that for a fixed number of inputs during training, the performance of PS-FCN increases with the number of inputs during testing. This is a desired property for photometric stereo as we can simply capture more images for robust estimation. 
For the rest of this paper, we refer PS-FCN as the model trained on both datasets and with an input of $32$ image-light pairs per sample.

\begin{table}[t] \centering
    \caption{\rev{Results of PS-FCN on {\sc Bunny} rendered using three different lighting distributions.}} \label{tab:res_synth_mirco_baseline}
    \includegraphics[width=0.09\textwidth]{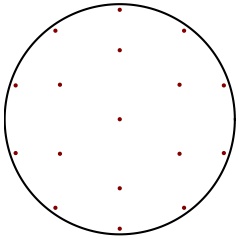}
\includegraphics[width=0.09\textwidth]{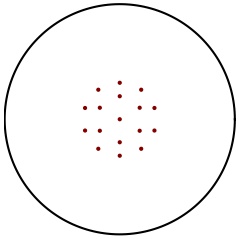}
\includegraphics[width=0.09\textwidth]{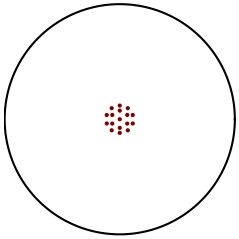}
\raisebox{0.9\height}{
\resizebox{0.19\textwidth}{!}{
\large
\begin{tabular}{ccc}
    \toprule
    Type & Range & MAE \\
    \midrule
    (a) & 144\degree$\times$144\degree & 4.21 \\ 
    (b) & 37\degree$\times$37\degree & 10.90 \\
    (c) & 22\degree$\times$22\degree & 18.72 \\
    \bottomrule
\end{tabular}
}}
\\
\vspace{-0.2em}
\makebox[0.09\textwidth]{\footnotesize (a)} 
\makebox[0.09\textwidth]{\footnotesize (b)}
\makebox[0.09\textwidth]{\footnotesize (c)}
\makebox[0.19\textwidth]{\footnotesize (d) Normal estimation}
\\

\end{table}

\begin{figure}[t] \centering
    \begin{subfigure}[t]{0.49\textwidth}
        \input{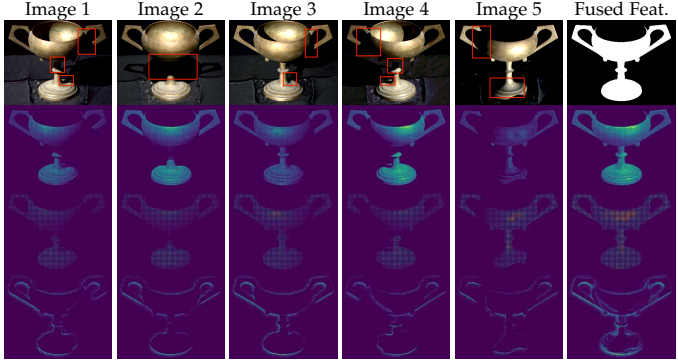}\vspace{-1.5em}
        \caption{The first five columns show the input images and the extracted features for each image (only $3$ out of $128$ feature channels are shown). The last column shows the object mask and the fused features by max-pooling. Red boxes in the images indicate regions with cast shadows.} \label{fig:visual_cast_shadow}
    \end{subfigure} \\ \vspace{1em}
    \begin{subfigure}[t]{0.49\textwidth}
        \input{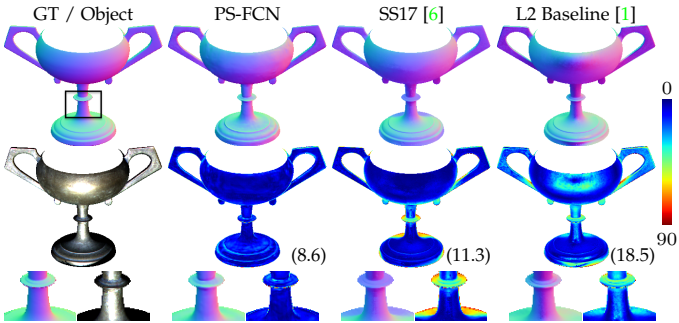}
        \caption{Comparison between PS-FCN, SS17~\cite{santo2017deep} and L2 Baseline~\cite{woodham1980ps} on {\sc Goblet}. The first row shows the ground-truth and estimated normals, and the second row shows the object and the error maps.} \label{fig:compare_cast_shadow} %
    \end{subfigure}
    \caption{Illustration of how max-pooling fusion layer handles surface regions with cast shadow using {\sc Goblet} from the DiLiGenT benchmark. (Note that the provided object mask and ground-truth normal map do not include the concave interior of {\sc Goblet}.)} \label{fig:cast_shadow}
\end{figure}

\begin{figure}[t] \centering
    \input{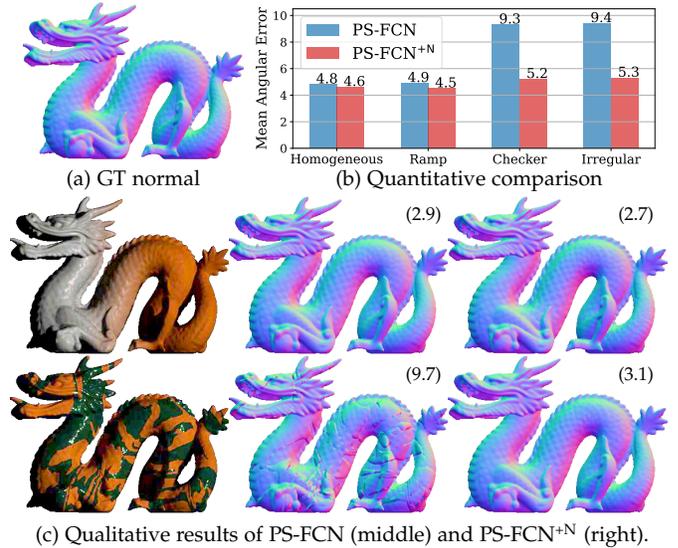}
    \caption{Comparison between PS-FCN and PS-FCN$^\text{+N}$ on {\sc dragon}$^{\text{SVBRDF}}$ dataset.} \label{fig:res_synth_SVBRDF} %
\end{figure}

\paragraph{Effects of lighting distributions} \rev{We tested PS-FCN on {\sc Bunny} rendered with three different lighting distributions, as shown in \Tref{tab:res_synth_mirco_baseline}. These three distributions have the same number of light source (\ie, $17$), but with different spanning ranges.
We can see that PS-FCN performs better when lightings are more diversely distributed. For the highly clustered distribution (see \Tref{tab:res_synth_mirco_baseline}~(c)), the results of PS-FCN drops notably. Since the lightings are randomly sampled from the upper-hemisphere (\ie, spanning range of $180\degree \times 180\degree$) during training, it is therefore not surprising to see PS-FCN with decreased performance under this extreme lighting distribution.
}

\paragraph{Results on surface with cast shadows} 
The presence of cast shadow is almost inevitable when the geometry of the object is non-convex, and is one of the major difficulties in photometric stereo.
Given the observation that a real surface point is unlikely to be shadowed under all light directions, we argue that max-pooling fusion can naturally overcome the effect of cast shadow when determining the surface normals.
This is because even a surface point is shadowed under some light directions, it can be observed under other light directions, and max-pooling can ignore those non-activated features and aggregate those activated features. 
\Fref{fig:cast_shadow}~(a) visualizes how max-pooling aggregates features from multiple observations and handles cast shadow. Compared with L2 baseline~\cite{woodham1980ps} and SS17~\cite{santo2017deep}, our method is more robust in regions with cast shadow (see \fref{fig:cast_shadow}~(b)).

\begin{figure*}[t] \centering
    \input{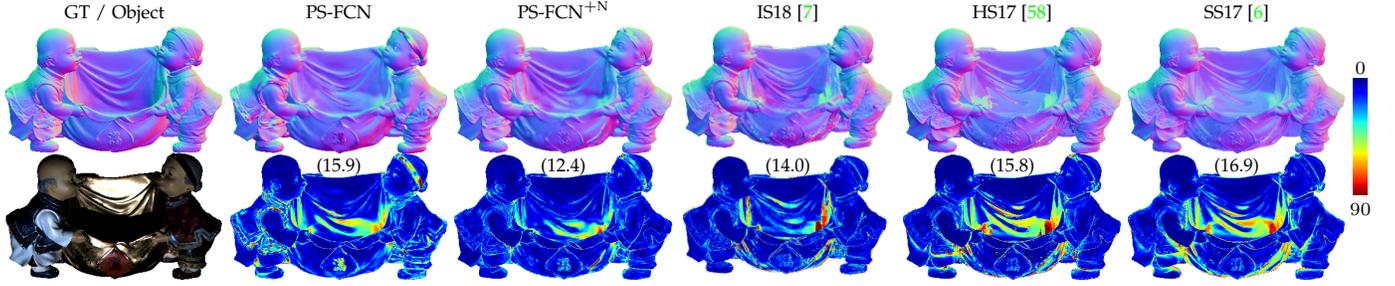} 
    \caption{Qualitative results on {\sc Harvest} in the DiLiGenT benchmark. Compared with PS-FCN, PS-FCN$^{+\text{N}}$ performs better for surfaces with SVBRDFs. In contrast with the per-pixel normal estimation methods~\cite{ikehata2018cnn,hui2017shape,santo2017deep}, PS-FCN$^{+\text{N}}$ can take advantage of the surface smooth prior and estimate a smoother normal map with less noise artifacts.} \label{fig:res_harvest}
\end{figure*}

\paragraph{Results on surfaces with SVBRDFs} 
To analyze how PS-FCN deteriorates in dealing with surfaces with SVBRDFs and verify the effectiveness of the proposed data normalization strategy, we compared PS-FCN and PS-FCN$^\text{+N}$ on {\sc dragon}$^{\text{SVBRDF}}$ dataset and the results are summarized in \fref{fig:res_synth_SVBRDF}. We can see that both models perform well on surfaces with homogeneous materials or surfaces with smooth BRDF changes (\eg, surfaces blended with Ramp). However, PS-FCN has difficulty in dealing with steep color changes caused by SVBRDFs (\eg, surfaces blended with Checker and Irregular). In contrast, PS-FCN$^\text{+N}$ is robust in handling surfaces with different types of SVBRDFs, which clearly demonstrates the effectiveness of the proposed data normalization strategy.

\begin{table}[t]
    \caption{Quantitative comparison of calibrated photometric stereo on the DiLiGenT benchmark.} \label{tab:quant_main}
    \centering
\resizebox{0.49\textwidth}{!}{
    \Huge
    \begin{tabular}{c|*{10}{c}|c}
        \toprule
        Method & {\sc ball}         & {\sc cat}          & {\sc pot1}         & {\sc bear}         & {\sc pot2}         & {\sc budd.}        & {\sc gobl.}         & {\sc read.}         & {\sc cow}            &{\sc harv.}        & Avg.\\
        \midrule
L2~\cite{woodham1980ps}   &\cl{4.1}     &\cl{8.4}     &\cl{8.9}     &\cl{8.4}     &\cl{14.7}    &\cl{14.9}    &\cl{18.5}    &\cl{19.8}     &\cl{25.6}    &\cl{30.6}     &\cl{15.4} \\
AZ08~\cite{alldrin2008p}  &\cl{2.7}     &\cl{6.5}     &\cl{7.2}     &\B{\cl{6.0}}     &\cl{11.0}    &\cl{12.5}    &\cl{13.9}    &\cl{14.2}     &\cl{21.5}    &\cl{30.5}     &\cl{12.6} \\
WG10~\cite{wu2010robust}  &\cl{2.1}     &\cl{6.7}     &\cl{7.2}     &\cl{6.5}     &\cl{13.1}    &\cl{10.9}    &\cl{15.7}    &\cl{15.4}     &\cl{25.9}    &\cl{30.0}     &\cl{13.4} \\
IA14~\cite{ikehata2014p}  &\cl{3.3}     &\cl{6.7}     &\cl{6.6}     &\cl{7.1}     &\cl{8.8}     &\cl{10.5}    &\cl{9.7}     &\cl{14.2}     &\cl{13.1}    &\cl{26.0}     &\cl{10.6} \\
ST14~\cite{shi2014bi}     &\cl{1.7}     &\cl{6.1}     &\cl{6.5}    &\cl{6.1}     &\cl{8.8}     &\cl{10.6}    &\cl{10.1}    &\cl{13.6}     &\cl{13.9}    &\cl{25.4}     &\cl{10.3} \\  
SS17~\cite{santo2017deep} &\cl{2.0}     &\cl{6.5}     &\cl{7.1}     &\cl{6.3}     &\cl{7.9}     &\cl{12.7}    &\cl{11.3}    &\cl{15.5}     &\cl{8.0}     &\cl{16.9}     &\cl{9.4} \\
TM18~\cite{Taniai18}      &\cl{1.5}     &\cl{5.4}     &\cl{6.1}    &\cl{5.8}     &\cl{7.8}     &\cl{10.4}    &\cl{11.5}  &\cl{11.0}  &\B{\cl{6.3}}   &\cl{22.6}  &\cl{8.8}   \\
HS17~\cite{hui2017shape}  &\B{\cl{1.3}} &\cl{4.9}     &\B{\cl{5.2}} &\B{\cl{5.6}}  &\cl{6.4}   &\cl{8.5}    &\cl{7.6}   &\cl{12.1}  &\cl{8.2}   &\cl{15.8}  &\cl{7.6}   \\
IS18$^*$~\cite{ikehata2018cnn} &\cl{2.2} &\B{\cl{4.6}} &\cl{5.4}    &\cl{8.3}  &\B{\cl{6.0}}   &\cl{7.9}   &\B{\cl{7.3}}   &\cl{12.6}  &\cl{8.0}   &\cl{14.0}  &\cl{7.6}  \\
    \midrule
PS-FCN                   &\cl{2.8}	 &\cl{6.2}	 &\cl{7.1}	 &\cl{7.6}	 &\cl{7.3}	 &\cl{7.9}	 &\cl{8.6}	 &\cl{13.3}	 &\cl{7.3}	 &\cl{15.9}	 &\cl{8.4} \\
PS-FCN$^\text{+N}$       &\cl{2.7}	 &\cl{4.8}	 &\cl{6.2}	 &\cl{7.7}	 &\cl{7.2}	 &\B{\cl{7.5}}	 &\cl{7.8}	 &\B{\cl{10.9}}	 &\cl{6.7}	 &\B{\cl{12.4}}	 &\B{\cl{7.4}} \\
        \bottomrule                                                                     
    \end{tabular}
}
 \\
    \begin{flushleft}
    {\scriptsize $^*$ indicates that the results of IS18~\cite{ikehata2018cnn} on {\sc bear} was computed using all of the $96$ images. The results reported in IS18~\cite{ikehata2018cnn} ({\sc Bear}: $4.1$, Avg.: $7.2$) was evaluated by discarding the first $20$ images. When discarding the first $20$ images, our results are PS-FCN ({\sc Bear}: $5.0$, Avg.: $8.1$) and PS-FCN$^\text{+N}$ ({\sc Bear}: $5.0$, Avg.: $7.1$).}
    \end{flushleft}
\end{table}

\subsubsection{Evaluation on real datasets}
We compared our method against the recently proposed learning based methods~\cite{santo2017deep,Taniai18,ikehata2018cnn} and other previous state-of-the-art methods on the DiLiGenT benchmark, as shown in \Tref{tab:quant_main}. 
After training with the data normalization strategy, PS-FCN$^{+\text{N}}$ performs better than PS-FCN on almost all of the ten objects, except for {\sc Bear}. 
Compared with the other state-of-the-art methods, PS-FCN$^{+\text{N}}$ performs particularly well on surface with complexed geometry and/or SVBRDFs~(\eg, {\sc buddha}, {\sc reading}, and {\sc harvest}), and achieves state-of-the-art results with an average MAE of $7.4$.
Qualitative comparison on {\sc Harvest} is shown in \fref{fig:res_harvest}.
\rev{Note that PS-FCN did not outperform previous methods on all the $10$ objects. We hypothesize that this might be caused by the limited training data. Different from pixel-wise approaches like IS18~\cite{ikehata2018cnn} and HS17~\cite{hui2017shape}, our method relies on diverse surface patches for training, while the current training data are only generated from $18$ objects.}

\subsection{Results on uncalibrated photometric stereo}

To measure the accuracy of the predicted light directions, the widely used mean angular error (MAE) in degree is adopted.
Since the light intensities among the testing images can only be estimated up to a scale factor $s$, we introduce the scale-invariant relative error (RE)
\begin{equation}
    RE_{scale} = \frac{1}{q} \sum_i^q \left(\frac{|s e_i -\tilde{e}_i|}{\tilde{e}_i} \right),
\end{equation}
where $q$ is the number of images, $e_i$ and $\tilde{e}_i$ are the estimated and ground-truth light intensities, respectively, for image $i$. The scale factor $s$ is computed by solving \hbox{$\argmin_s \sum_i^n (s e_i -\tilde{e}_i)^2$} with least squares. 

\subsubsection{Network analysis of LCNet with synthetic data}
\begin{figure}[t] \centering
    \input{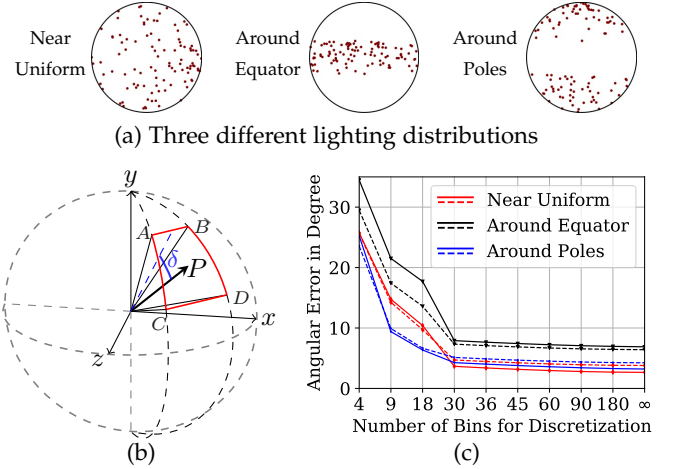}
    \caption{(a) Three different lighting distributions. (b) Light directions $A$, $B$, $C$, and $D$ have the maximum deviation angles with the actual light direction $P$ after discretization. (c) Normal estimation error of PS-FCN on {\sc Sphere} (solid lines) and {\sc Bunny} (dashed lines) under different light direction space discretization levels ($\infty$ indicates no discretization).} \label{fig:discretization}
\end{figure}

For all experiments on synthetic dataset involving input with unknown light intensities, we randomly generated light intensities in the range of $[0.2, 2.0]$. Each experiment was repeated five times and the average results were reported.

\paragraph{Discretization of lighting space}
For a given number of bins $K_d$, the maximum deviation angle for azimuth and elevation of a light direction is $\delta = 180\degree/(K_d\times 2)$ after discretization (\eg, $\delta=2.5\degree$ when $K_d=36$).
\revision{Note that discretizing azimuth and elevation angles independently indicates that lighting space is more densely discretized around the poles and less around the equator. 
This suggests that the link between the quantization of the lighting space and surface normal estimation error correlates with the lighting distribution.
To investigate how the light direction discretization affects the surface normal estimation accuracy, we tested PS-FCN on {\sc Sphere} and {\sc Bunny} rendered under three different lighting distributions, namely, \textit{Near Uniform}, \textit{Around Equator}, and \textit{Around Poles} (see \fref{fig:discretization}~(a)).}

We divided the azimuth and elevation angles of light directions into different numbers of bins ranging from $4$ to $180$.
For a specific bin number, we perturbed the azimuth and elevation of each ground-truth light direction by the maximum deviation angle, leading to four light directions that have the maximum possible angular deviations after discretization (see \fref{fig:discretization}~(b)).
We then used these light directions as input for PS-FCN to infer surface normals. The normal estimation error reported in \fref{fig:discretization} (c) is the upper-bound error for PS-FCN caused by discretization.
\rev{We can see that the error increase caused by discretization is marginal for all three lighting distributions when $K_d\ge30$.}
We chose a relatively sparse discretization of lighting space in this paper as it allows PS-FCN to learn to better tolerate small errors in the estimated lighting at test time.

\newcommand{\LCNetreg}{LCNet$_\text{reg}$\xspace}
\begin{table}[t] \centering
    \caption{Lighting estimation results (MAE in degree for light direction and relative error for intensity) on SynTest$^\text{MERL}$ dataset. The results are averaged over samples rendered with $100$ BRDFs. (Value the lower the better)}
    \resizebox{0.49\textwidth}{!}{
    \large
\begin{tabular}{l|*{2}{c}|*{2}{c}|*{2}{c}|*{2}{c}}
    \toprule
    & \multicolumn{2}{c}{{\sc Sphere}} & \multicolumn{2}{c}{{\sc Bunny}} & \multicolumn{2}{c}{{\sc Dragon}} & \multicolumn{2}{c}{{\sc Armadillo}}\\
    Model  & Dir.    & Int.  &  Dir.    & Int.  &  Dir.    & Int. &  Dir.    & Int.\\
    \midrule
    LCNet                     & \B{3.47} & \B{0.082} & \B{5.38}  & \B{0.089} & \B{7.85} & \B{0.096} & \B{7.50}  & \B{0.103}\\ %
    LCNet$_{\text{w/o mask}}$ & 5.46     & 0.104     & 8.85      & 0.144 & 11.81 & 0.176 & 13.02 & 0.166\\ %
    LCNet$_{\text{local}}$    & 6.87     & 0.198     & 9.98      & 0.255 & 10.58 & 0.264 & 9.50 & 0.266 \\ %
    \bottomrule
\end{tabular}
}
 \label{tab:quant_light_synth}
\end{table}
\begin{table}[t] \centering
    \caption{Results of LCNet and \LCNetreg on {\sc Sphere} and {\sc Bunny} rendered under different lighting distributions.} \label{tab:quant_light_synth_regression}
    \resizebox{0.49\textwidth}{!}{
    \Large
\begin{tabular}{l|*{4}{c}|*{4}{c}|*{4}{c}}
    \toprule
    & \multicolumn{4}{c}{Near Uniform} & \multicolumn{4}{c}{Around Equator} & \multicolumn{4}{c}{Around Poles} \\
    & \multicolumn{2}{c}{{\sc Sphere}} & \multicolumn{2}{c|}{{\sc Bunny}} & \multicolumn{2}{c}{{\sc Sphere}} & \multicolumn{2}{c}{{\sc Bunny}} & \multicolumn{2}{c}{{\sc Sphere}} & \multicolumn{2}{c}{{\sc Bunny}}\\
    Model  & Dir.    & Int.  &  Dir.    & Int.  &  Dir.    & Int. &  Dir.    & Int. &  Dir.    & Int. &  Dir.    & Int. \\
    \midrule
    LCNet                 & \B{3.47} & \B{0.082} & \B{5.38}  & \B{0.089} & \B{3.32} & \B{0.079} & \B{5.33} & \B{0.077} & \B{4.82} & \B{0.088} & \B{6.34} & \B{0.095} \\
    \LCNetreg  & 4.10     & 0.104     & 5.46  & 0.094  & 3.72 & 0.091 & 5.85 & 0.092 & 5.57 & 0.104 & 7.47 & 0.102 \\ 
    \bottomrule
\end{tabular}
}
 
\end{table}

\paragraph{Effectiveness of LCNet}
\rev{We first investigated the effect of object mask input and local-global feature fusion.
\Tref{tab:quant_light_synth} shows that taking the object mask as input and adopting the proposed local-global feature fusion strategy can effectively improve the lighting estimation results.}

\rev{We then compared LCNet with a regression based baseline, denoted as \LCNetreg, to validate the effectiveness of the classification based model (please refer to the supplementary material for implementation details). Specifically, we tested LCNet and \LCNetreg on three different lighting distributions 
illustrated in \fref{fig:discretization}~(a). The results are shown in \Tref{tab:quant_light_synth_regression}. The proposed classification based LCNet consistently outperforms \LCNetreg on both light direction and intensity estimation. This echoes our hypothesis that classifying a light direction to a certain range is easier than regressing an exact value. Thus, solving the classification problem reduces the learning difficulty and improves the performance.}
\revision{It can also be seen that both methods perform better on \textit{Around Equator} and worse on \textit{Around Poles}.
This suggests that lightings around the poles are more difficult to estimate due to their extremely directions, independent of the lighting space discretization.}

\begin{figure}[t] \centering
    \includegraphics[width=0.235\textwidth]{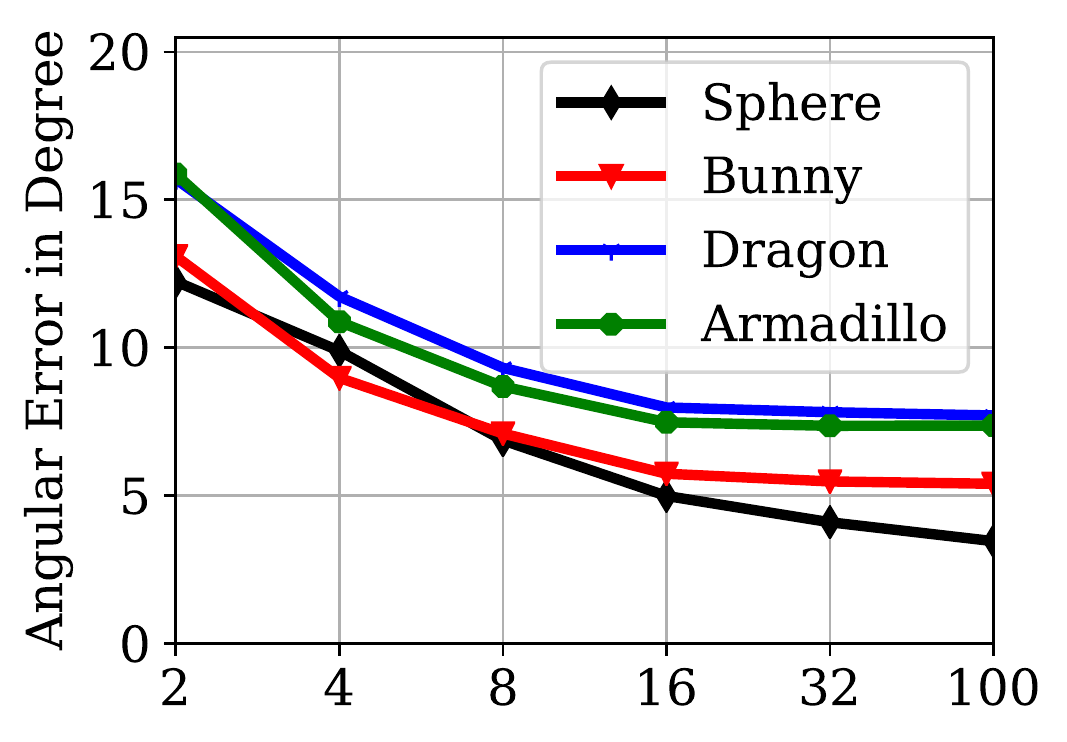}
    \includegraphics[width=0.245\textwidth]{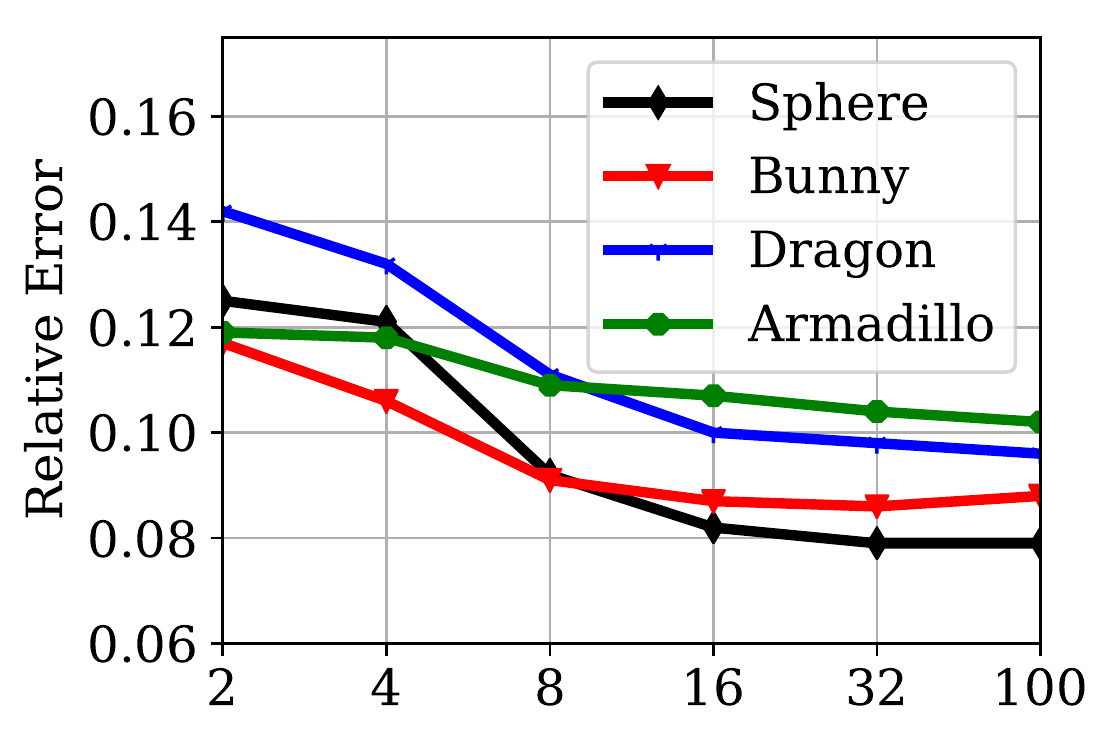}\\ \vspace{-0.5em}
    \makebox[0.235\textwidth]{\scriptsize (a) Light direction estimation}\hfill
    \makebox[0.245\textwidth]{\scriptsize (b) Light intensity estimation}\\
    \caption{Lighting estimation results of LCNet on SynTest$^\text{MERL}$ dataset with varying input image numbers.} \label{fig:img_num_syn}
\end{figure}

\begin{table}[t] \centering
    \caption{Lighting estimation results of LCNet on {\sc dragon}$^{\text{SVBRDF}}$ dataset.}
    \resizebox{0.47\textwidth}{!}{
    \large
\begin{tabular}{l|*{2}{c}|*{2}{c}|*{2}{c}|*{2}{c}}
    \toprule
    & \multicolumn{2}{c}{Homogeneous} & \multicolumn{2}{c}{Ramp} & \multicolumn{2}{c}{Checker} & \multicolumn{2}{c}{Irregular}\\
     \makebox[0.01\textwidth]{} Model  \makebox[0.01\textwidth]{} & Dir. & Int. &  Dir. & Int.  &  Dir.    & Int. &  Dir.    & Int.\\
    \midrule
    \makebox[0.01\textwidth]{} LCNet \makebox[0.01\textwidth]{}& 7.85 & 0.096 & 7.44  & 0.076 & 8.19 & 0.111 & 8.68  & 0.103 \\
    \bottomrule
\end{tabular}
}
 \label{tab:quant_light_synth_SVBRDF}
\end{table}

\begin{table}[t] \centering
    \caption{Normal estimation results on SynTest$^\text{MERL}$ dataset. PS-FCN$^{\dag}$ was trained given lightings estimated by LCNet or \LCNetreg.} %
    \resizebox{0.49\textwidth}{!}{
\Huge
\begin{tabular}{cl|c|*{1}{c}*{1}{c}*{1}{c}*{1}{c}}
\toprule
ID & Model  & \# Param & {\sc Sphere} & {\sc Bunny} & {\sc Dragon} & {\sc Armad.}\\
\midrule
C0 & PS-FCN                      & 2.2 M & 2.66   & 3.80 & 4.83 & 5.24\\ %
\midrule \rowcolor{gray!30}
C1 & LCNet + PS-FCN$^\dag$& 6.6 M        & \B{2.71} & \B{4.09} & \B{6.41} & \B{7.09}\\
C2 & LCNet + PS-FCN                      & 6.6 M & 3.19  & 4.67 & 6.92 & 7.70\\ %
C3 & LCNet$_{\text{reg}}$ + PS-FCN$^\dag$ & 6.6 M & 3.22  & 4.99 & 6.63 & 7.54\\ %
C4 & LCNet + PS-FCN$^{+\text{N}}$        & 6.6 M & 4.53  & 5.35 & 7.36 & 7.99\\ %
\midrule
C5 & UPS-FCN$_{\text{deep+mask}}$   & 6.1 M & 3.65 & 6.41 & 9.68 & 11.26   \\ %
C6 & UPS-FCN     & 2.2 M & 7.44 & 12.34 & 14.44 & 15.93 \\ %
\bottomrule
\end{tabular}
}
 \label{tab:quant_normal_syn}
\end{table}

\Fref{fig:img_num_syn} shows that the performance of LCNet increases with the number of input images. This is expected, since more useful information can be used to infer lightings with more input images.

To analyze the effect of SVBRDFs in lighting estimation, we tested LCNet on {\sc dragon}$^{\text{SVBRDF}}$ dataset and reported the result in \Tref{tab:quant_light_synth_SVBRDF}.
We can see that LCNet is robust to surfaces with SVBRDFs, which indicates that unlike surface normal estimation, the features used by LCNet for lighting estimation are robust to SVBRDFs (\eg, attached shadow).

\paragraph{Integration with PS-FCN}
For uncalibrated photometric stereo, we train a variant of PS-FCN, denoted as PS-FCN$^\dag$, using the lighting estimated by LCNet. Note that the weights of LCNet was fixed during the training of PS-FCN$^\dag$, as we found that end-to-end fine-tuning did not improve the performance.  
Experiments with IDs C1 \& C2 in \Tref{tab:quant_normal_syn} show that after training with the discretized lighting estimated by LCNet, PS-FCN$^\dag$ performs better than PS-FCN given possibly noisy lightings at test time. Besides, experiments with IDs C1 \& C3 show that PS-FCN$^\dag$ coupled with the classification based LCNet consistently outperforms that with the regression based \LCNetreg. 

\rev{However, it is not straightforward to integrate LCNet with PS-FCN$^{+\text{N}}$ to handle surface with SVBRDFs. Experiments with IDs C2 \& C4 show that integrating LCNet with PS-FCN$^{+\text{N}}$ decreases the performance of normal estimation.}
This is because when the estimated light intensities are noisy, the data normalization operation will magnify this error. We experimentally found that retraining PS-FCN$^{+\text{N}}$ with the estimated lighting of LCNet cannot improve the result.
\rev{In the future, we will investigate better strategies for handling surfaces with SVBRDFs under the uncalibrated setup.}

\begin{figure}[t]\centering
    \input{figures/compare_quant_pf14_v2}
    \caption{Comparison between LCNet+PS-FCN$^\dag$ and PF14~\cite{papad14closed} on {\sc Bunny} rendered with four different types of BRDFs under a near uniform lighting distribution and a biased lighting distribution.} \label{fig:compare_pf14}
\end{figure}

\paragraph{Comparison with single-stage models}
To validate the effectiveness of the proposed two-stage framework, we compared our method with two different single-stage baseline models.
We first train a variant of PS-FCN, denoted as UPS-FCN, without taking the light direction as input during training and testing.
We then increased the model capacity of UPS-FCN by training a deeper network, denoted as UPS-FCN$_{\text{deep+mask}}$, that takes both the images and object mask as input. 
Please refer to our supplementary material for detailed network architectures.

Experiments with IDs C5 \& C6 in \Tref{tab:quant_normal_syn} show that utilizing a deeper network and taking the object mask as input can improve the performance of single-stage model.
However, experiments with IDs C1 \& C5 show that the proposed method significantly outperforms the single-stage model, when the input as well as the number of parameters are comparable.
This result indicates that simply increasing the depth of the network cannot produce optimal results. 

\paragraph{Comparison with the non-learning method~\cite{papad14closed}}
To further verify the effectiveness of our method over non-learning method, we compared our method with the existing uncalibrated method PF14~\cite{papad14closed}, which achieves state-of-the-art results on the DiLiGenT benchmark~\cite{shi2018benchmark}.
\rev{\Frefs{fig:compare_pf14}~(c)-(d) show that our method consistently outperforms PF14 on {\sc Bunny} rendered using different types of BRDFs (\ie, Lambertian, Fabric, Plastic, and Phenolic) under two different lighting distributions (one near uniform and one biased), especially on surfaces exhibiting specular highlights.}

\begin{table}[t] \centering
    \caption{Quantitative results on the DiLiGenT benchmark.} \label{tab:quant_light_normal_diligent}
    \makebox[0.49\textwidth]{(a) Results on light direction estimation.}
    \resizebox{0.49\textwidth}{!}{
            \huge
	         \begin{tabular}{l|*{10}{c}|c}
            \toprule
 Method           & {\sc ball}         & {\sc cat}          & {\sc pot1}         & {\sc bear}         & {\sc pot2}         & {\sc budd.}        & {\sc gobl.}         &{\sc read.}         & {\sc cow}            &{\sc harv.}        & Avg.\\
            \midrule
            LCNet$_{\text{reg}}$ & 4.94	 & 5.82	 & 5.62	 & 7.19	 & 4.82	 & \B{3.90}	 & 12.89	 & 7.90	 & \B{4.19}	 & 9.50	 & 6.68 \\
            LCNet & \B{3.27}	 & \B{4.08}	 & \B{5.44}	 & \B{3.47}	 & \B{2.87}	 & 4.34	 & \B{10.36}	 & \B{4.50}	 & 4.52	 & \B{6.32}	 & \B{4.92} \\
	             \bottomrule
        \end{tabular}
}	 

 \vspace{0.4em}
    \makebox[0.49\textwidth]{(b) Results on light intensity estimation.}
    \resizebox{0.49\textwidth}{!}{
            \huge
	         \begin{tabular}{l|*{10}{c}|c}
            \toprule
 Method           & {\sc ball}         & {\sc cat}          & {\sc pot1}         & {\sc bear}         & {\sc pot2}         & {\sc budd.}        & {\sc gobl.}         &{\sc read.}         & {\sc cow}            &{\sc harv.}        & Avg.\\
            \midrule
            LCNet$_{\text{reg}}$& \B{0.032}	 & \B{0.051}	 & \B{0.048}	 & 0.167	 & 0.074	 & 0.080	 & 0.075	 & 0.141	 & \B{0.044}	 & 0.085	 & 0.080 \\
            LCNet & 0.039	 & 0.095	 & 0.058	 & \B{0.061}	 & \B{0.048}	 & \B{0.048}	 & \B{0.067}	 & \B{0.105}	 & 0.073	 & \B{0.082}	 & \B{0.068} \\
	             \bottomrule
        \end{tabular}
}	 

 \vspace{0.4em}
    \makebox[0.49\textwidth]{(c) Results on normal estimation.}
    \resizebox{0.49\textwidth}{!}{
    \Huge
\begin{tabular}{l|*{10}{c}|c}
    \toprule
    Method & {\sc ball}         & {\sc cat}          & {\sc pot1}         & {\sc bear}         & {\sc pot2}         & {\sc budd.}        & {\sc gobl.}         & {\sc read.}         & {\sc cow}            &{\sc harv.}        & Avg.\\
    \midrule
    AM07~\cite{alldrin2007r}      & \cl{7.3} & \cl{31.5} & \cl{18.4} &  \cl{16.8} & \cl{49.2} & \cl{32.8} & \cl{46.5} & \cl{53.7}& \cl{54.7} & \cl{61.7} & \cl{37.3} \\
    SM10~\cite{shi2010self}       & \cl{8.9} & \cl{19.8} & \cl{16.7} &  \cl{12.0} & \cl{50.7} & \cl{15.5} & \cl{48.8} & \cl{26.9}& \cl{22.7} & \cl{73.9} & \cl{29.6} \\
    WT13~\cite{wu2013calib}       & \cl{4.4} & \cl{36.6} & \cl{9.4}  &  \textbf{\cl{6.4}}  & \cl{14.5} & \cl{13.2} & \cl{20.6} & \cl{59.0}& \cl{19.8} & \cl{55.5} & \cl{23.9} \\
    LM13~\cite{lu2013uncalibrated}& \cl{22.4}& \cl{25.0} & \cl{32.8} &  \cl{15.4} & \cl{20.6} & \cl{25.8} & \cl{29.2} & \cl{48.2}& \cl{22.5} & \cl{34.5} & \cl{27.6} \\
    PF14~\cite{papad14closed}     & \cl{4.8} & \cl{9.5}  & \cl{9.5}  &  \cl{9.1}  & \cl{15.9} & \cl{14.9} & \cl{29.9} & \cl{24.2}& \cl{19.5} & \cl{29.2} & \cl{16.7} \\
    LC18~\cite{lu2018symps}       &\cl{9.3}  &  \cl{12.6} &\cl{12.4} &\cl{10.9}	 &\cl{15.7} &\cl{19.0} &\cl{18.3} &\cl{22.3} &\cl{15.0} &\cl{28.0} &\cl{16.3} \\
    \midrule
    UPS-FCN$_{\text{deep+mask}}$  &\cl{4.0}	 &\cl{12.2}	 &\cl{11.1}	 &\cl{7.2}	 &\cl{11.1}	 &\cl{13.1}	 &\cl{18.1}	 &\cl{20.5}	 &\cl{11.8}	 &\cl{27.2}	 &\cl{13.6} \\
    LCNet$_{\text{reg}}$+PS-FCN$^\dag$ &\cl{3.9}	 &\cl{9.00}	 &\textbf{\cl{8.0}}	 &\cl{16.0}	 &\cl{8.4}	 &\cl{9.4}	 &\textbf{\cl{11.5}}	 &\cl{17.0}	 &\cl{8.8}	 &\cl{18.4}	 &\cl{11.0} \\
    LCNet+PS-FCN$^\dag$ &\textbf{\cl{2.8}}	 &\textbf{\cl{8.1}}	 &\textbf{\cl{8.1}}	 &\cl{6.9}	 &\textbf{\cl{7.5}}	 &\textbf{\cl{9.00}}	 &\textbf{\cl{11.9}}	 &\textbf{\cl{14.9}}	 &\textbf{\cl{8.5}}	 &\textbf{\cl{17.4}}	 &\textbf{\cl{9.5}} \\
    \bottomrule
\end{tabular}
}

\end{table}

\subsubsection{Evaluation on real datasets}
\Tref{tab:quant_light_normal_diligent} (a)-(b) show that LCNet outperforms the regression based baseline \LCNetreg and achieves highly accurate results on both light direction and intensity estimation on DiLiGenT benchmark, with an average MAE of $4.92$ and an average relative error of $0.068$, respectively.
\Tref{tab:quant_light_normal_diligent} (c) compares the normal estimation results of LCNet+PS-FCN$^\dag$ with previous state-of-the-art methods on DiLiGenT benchmark.
LCNet+PS-FCN$^\dag$ achieves state-of-the-art results on almost all objects with an average MAE of $9.5$, except for {\sc Bear}. 
Please refer to our supplementary material for more qualitative comparisons.

\subsubsection{LCNet and the GBR ambiguity}
\begin{figure}[t] \centering
    \input{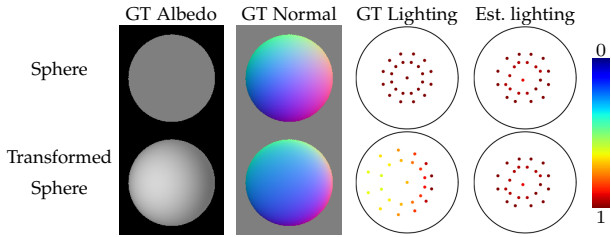}
    \caption{Results of LCNet on surfaces differed by GBR transformations. The GT albedo, normal map, and lightings in the first and second rows give the same input images.}
    \label{fig:sphere_gbr}
\end{figure}

\rev{\Eref{eq:gbr} in \sref{sec:related_work} indicates that when lightings are unknown, theoretically, the surface normal for a Lambertian surface can only be estimated up to the GBR transformation. This indicates that multiple combinations of albedo, normal, and lightings can result in the same set of images.  
However, the albedo of a GBR transformed surface becomes smoothly changing according to the surface normal, and such a spatially-varying albedo distribution seldom exists in the real-world~\cite{belhumeur1999bas}. 
\Fref{fig:sphere_gbr} shows that LCNet estimates the same lightings for two surfaces differed by GBR transformations, since the input images are the same. The estimated lightings are very close to lightings that correspond to the shape with uniform albedo (top row).
This result suggests that LCNet is trained to predict the most probable lightings by implicitly assuming the albedo distribution is not GBR transformed, which is similar to previous non-learning methods relying on albedo distribution to resolve the GBR ambiguity~\cite{alldrin2007r,shi2010self,papad14closed}.
As LCNet was trained using realistic surfaces (\ie, surfaces without GBR transformed albedo distributions), it tries to predict realistic surfaces that are learned rather than other possible but unrealistic surfaces, which are not included in the training set (bottom row).
}

\begin{figure}[t] \centering
    \input{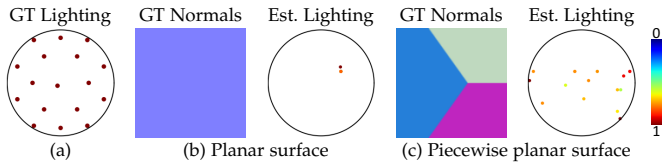} 
    \caption{Lighting estimation results of LCNet on a planar and a piecewise planar surfaces with uniform albedo.}\label{fig:quant_light_plane}
\end{figure}

\rev{
    \Fref{fig:quant_light_plane} shows that LCNet fails to estimate reliable lightings on two special ambiguous cases, where a planar and a piecewise planar surfaces are rendered with uniform albedos. For planar surface (\ie, the surface normal is constant over the whole surface), the ambiguity cannot be resolved even when assuming uniform albedos.
    Although there are three different surface normals in the piecewise planar surface, the surface integrability constraint~\cite{belhumeur1999bas} cannot be applied to reduce the linear ambiguity to GBR, making this case unsolvable as well. Note that existing methods also fail to handle these two ambiguous cases.
}

\section{Conclusions}
In this paper, we proposed a deep fully convolutional network, named PS-FCN, for calibrated photometric stereo. PS-FCN can accept an arbitrary number of images and their associated light directions as input and estimate an accurate normal map in a fast feed-forward pass. 
We then introduced a simple yet effective data normalization strategy to allow PS-FCN better deal with surfaces with spatially-varying BRDFs.
To handle the uncalibrated scenario, we introduced a convolutional network, named LCNet, to estimate lightings from input images. The estimated lightings and the input images can then be utilized by PS-FCN to estimate the surface normals.
Our method does not require a pre-defined set of light directions during training and testing. It can handle multiple images and light directions in an order-agnostic manner.  
In order to train our model, two large-scale synthetic datasets with various realistic shapes and materials have been created. After training, our model can generalize well on challenging real datasets.
Extensive results on both synthetic and real datasets have clearly shown that our method outperforms previous state-of-the-art methods on both calibrated and uncalibrated photometric stereo. 

\ifCLASSOPTIONcompsoc

\section*{Acknowledgments}
Kai Han is supported by EPSRC Programme Grant Seebibyte EP/M013774/1.
Boxin Shi is supported by National Natural Science Foundation of China under Grant No. 61872012, National Key R\&D Program of China (2019YFF0302902), and Beijing Academy of Artificial Intelligence (BAAI).
Yasuyuki Matsushita is supported by the New Energy and Industrial Technology Development Organization (NEDO).
Kwan-Yee~K.~Wong is supported by a grant from the Research Grant Council of the Hong Kong (SAR), China, under the project HKU 17203119.
\else
  \section*{Acknowledgment}
\fi

\ifCLASSOPTIONcaptionsoff
  \newpage
\fi

\bibliographystyle{IEEEtran}
\bibliography{IEEEabrv,gychen}

\end{document}